\documentclass[11pt,a4paper]{article}
\usepackage{times,latexsym}
\usepackage{url}
\usepackage[T1]{fontenc}

\usepackage[acceptedWithA]{tacl2021v1}

\usepackage{xspace,mfirstuc,tabulary}

\usepackage[dvipsnames]{xcolor}
\usepackage{booktabs}
\usepackage[table]{xcolor}
\usepackage{soul}
\usepackage{tcolorbox}
\usepackage{rotating}
\usepackage{booktabs}
\usepackage{multirow}
\tcbuselibrary{skins}
\usepackage{amsmath}
\usepackage{amssymb}

\definecolor{revcolor}{HTML}{C1440E}

\definecolor{revcolorb}{HTML}{0072B2}

\definecolor{revcolorg}{HTML}{2da02c}

\definecolor{boston_blue}{HTML}{3583bb}
\definecolor{forest_green}{HTML}{2da02c}
\definecolor{salmon}{HTML}{fa8072}
\definecolor{yang}{HTML}{2da02c}

\definecolor{AC}{RGB}{51,34,136}
\definecolor{AU}{RGB}{17,119,51}
\definecolor{TNG}{RGB}{68,170,153}
\definecolor{IE}{RGB}{136,204,238}
\definecolor{ST}{RGB}{221,204,119}
\definecolor{QU}{RGB}{204,102,119}
\definecolor{NE}{RGB}{170,68,153}
\definecolor{NS}{RGB}{136,34,85}
\definecolor{YW}{RGB}{255,204,204}
\definecolor{WP}{RGB}{229,204,255}

\definecolor{high}{HTML}{ec462e}
\definecolor{low}{HTML}{ffffff}
\newcommand*{\opacity}{90}
\newcommand*{\MinNumber}{0.0}
\newcommand*{\MidNumber}{7.0}
\newcommand*{\MaxNumber}{15.0}

\definecolor{purpleGreek}{HTML}{800080}
\definecolor{blueGreek}{HTML}{4891e7}
\definecolor{redGreek}{HTML}{d9381e}

\newcommand{\uom}{{\color{purpleGreek}\boldsymbol{\alpha}}}
\newcommand{\uos}{{\color{blueGreek}\boldsymbol{\beta}}}
\newcommand{\sufe}{{\color{redGreek}\boldsymbol{\gamma}}}

\newcommand{\colorcell}[1]{
    \pgfmathsetmacro{\value}{#1}

    \pgfmathparse{\value > \MaxNumber ? 1 : 0}
    \ifnum\pgfmathresult=1
        \hspace{-0.33em}\cellcolor{high!\opacity}{#1}
    \else
        \pgfmathparse{\value > \MidNumber ? 1 : 0}
        \ifnum\pgfmathresult=1
            \pgfmathparse{int(round(100*(\value - \MidNumber)/(\MaxNumber - \MidNumber)))}
            \xdef\tempa{\pgfmathresult}
            \hspace{-0.33em}\cellcolor{high!\tempa!yellow!\opacity}{#1}
        \else
            \pgfmathparse{int(round(100*(\MidNumber - \value)/(\MidNumber - \MinNumber)))}
            \xdef\tempa{\pgfmathresult}
            \hspace{-0.33em}\cellcolor{low!\tempa!yellow!\opacity}{#1}
        \fi
    \fi
}

\usepackage{silence}
\newif\iftaclinstructions
\taclinstructionsfalse
\iftaclinstructions
\renewcommand{\confidential}{}
\renewcommand{\anonsubtext}{(No author info supplied here, for consistency with
TACL-submission anonymization requirements)}
\newcommand{\instr}
\fi

\iftaclpubformat

\else

\fi

\title{
Tracing Knowledge and Reasoning in LLMs
}

\author{
  \parbox{\textwidth}{\centering
    \textbf{Yang Wang}$\hspace{.1em}^{\uom}$$\hspace{.1em}^{\dagger}$,
    \textbf{Xiao Wang}$\hspace{.1em}^{\uom}$$\hspace{.1em}^{\dagger}$,
    \textbf{Yiqi Liu}$\hspace{.1em}^{\uom}$$\hspace{.1em}^{\dagger}$, \\
    \vspace{0.1cm}
    \textbf{Chenghao Xiao}$\hspace{.1em}^{\sufe}$,
    \textbf{Eleni Vasilaki}$\hspace{.1em}^{\uos}$,
    \textbf{Goran Nenadic}$\hspace{.1em}^{\uom}$,
    \textbf{Chenghua Lin}$\hspace{.1em}^{\uom}$$\hspace{.1em}^{\ddagger}$ \\
    \vspace{0.4cm}
    \textnormal{$^{\uom}\hspace{.1em}$University of Manchester~~~}
    \textnormal{$^{\uos}\hspace{.1em}$University of Sheffield}\\
    \textnormal{$^{\sufe}\hspace{.1em}$Shanghai University of Finance and Economics}\\
    \vspace{0.2cm}
    \textnormal{\texttt{\{yang.wang-2,\,xiao.wang-4,\,chenghua.lin\}@manchester.ac.uk}}
  }
}

\date{}

\begin{document}
\maketitle

\begin{NoHyper}
\renewcommand{\thefootnote}{}
\footnotetext{$^{\dagger}$\hspace{.2em}Equal contribution.}
\footnotetext{$^{\ddagger}$\hspace{.2em}Corresponding author.}
\end{NoHyper}

\begin{abstract}
Current evaluations of Large Language Models (LLMs) report performance at the dataset or task level, revealing what a model achieves but offering little insight into the internal processing associated with its performance. We propose a task-centric framework for characterising the model-internal processing elicited by a benchmark. It combines attention- and activation-based signals with representation-based measures to characterise benchmarks spanning a continuum from retrieval-dominant to reasoning-dominant and to surface failure modes that task-level scores conceal. We evaluate 16 LLMs across 24 diverse benchmarks and conduct targeted internal-state analyses on representative models and tasks, observing systematic differences in processing between knowledge-centric and reasoning-centric tasks. We shift evaluation from asking what score a model achieves toward characterising the processing patterns associated with that score, potentially supporting more targeted benchmark selection and more informed model choice. We open source our code at \url{https://github.com/extraordinarylab/trace-kr}.
\end{abstract}

\section{Introduction}

Large Language Models (LLMs) have demonstrated remarkable capabilities across a diverse spectrum of tasks, achieving state-of-the-art results on numerous benchmarks \cite{wang2024mmlu, jimenez2024swebench, du2026supergpqa}. This progress has been largely measured by aggregate performance scores on static leaderboards. While invaluable for tracking progress, this evaluation paradigm is reaching its limits \cite{center2026benchmark, shi-etal-2026-judgeagent}.
A single accuracy number on a benchmark tells us what a model can do, but reveals little about how it achieves the result or why it fails, because it cannot tell whether a correct answer reflects memorised knowledge or actual reasoning.
As models become more powerful, simply measuring their aggregate success yields diminishing insight into their underlying architectural strengths and weaknesses. The field requires a more fine-grained approach to evaluation that can disentangle the fundamental capabilities underpinning performance.

The challenge of disentangling a model's knowledge from its reasoning is widely acknowledged in the community \cite{jin-etal-2025-disentangling-memory, wu2025knowledge}. Prior work has largely focused on analysing these capabilities in isolation, which has significant limitations.
Research into knowledge boundaries (i.e., what an LLM knows versus what it does not) primarily uses techniques like uncertainty estimation, confidence calibration, and internal state probing \cite{li-etal-2025-knowledge-boundary, xiao-etal-2025-analyzing}.
Recent work nevertheless shows that reliable boundary estimation remains difficult: uncertainty can be imperfectly calibrated, while poor awareness of parametric knowledge boundaries can lead to inappropriate reliance on internal or retrieved information \cite{ni-etal-2025-towards, li-etal-2026-demystifying, feng-etal-2026-trust}.
Studies of mathematical and compositional reasoning reveal that performance can degrade under seemingly superficial perturbations \cite{mirzadeh2025gsmsymbolic}. More recent evidence further shows that models may possess the factual information needed to solve a task yet still rely on surface heuristics rather than applying that knowledge through principled reasoning \cite{lehmann-etal-2026-knowing}.
While valuable, these largely isolated approaches leave open how knowledge retrieval and reasoning interact within hybrid tasks. Recent work \cite{feng-etal-2026-tracking} has begun to study this interaction, showing that failures can arise either from inadequate knowledge integration or from downstream reasoning even when the relevant knowledge is available.
A high score on a ``reasoning benchmark'' does not reveal whether success comes from robust reasoning or from easier retrieval shortcuts. For example, a science benchmark may contain questions that require recalling a specific fact before applying it in a multi-step inference; strong aggregate performance does not show which of these two capabilities the model actually relies on.
Recent evidence further suggests that access to relevant knowledge does not guarantee its effective use: models may attend to shortcut-like cues or fall back on parametric memory even when sufficient external knowledge is provided \cite{wu2025knowledge, li-etal-2026-llms-hallucinate}.

Recent studies have begun to address this gap by moving beyond passive observation toward active decoupling of these capabilities. For instance, some frameworks propose decomposing the inference process into explicit steps by fine-tuning models with special tokens like \texttt{<memory>} and \texttt{<reason>} \cite{jin-etal-2025-disentangling-memory}, or by prompting models to adopt \textit{fast} (knowledge-based) versus \textit{slow} (reasoning-based) thinking modes inspired by dual-system theory \cite{yang2025decoupling}. Another line of work focuses on evaluating the final reasoning chain, introducing metrics to judge the correctness of knowledge and the logical quality of each step \cite{wu2025knowledge}.
Although effective in their respective settings, these methods either alter the model's natural inference process or rely solely on analysing the generated text. Consequently, they do not provide a task-level account of the relative knowledge-retrieval and reasoning demands reflected in the internal processing of an unmodified LLM.

To address this limitation, we draw on prior findings concerning the internal structure of Transformers.
Prior work \cite{yang2025decoupling, song2025demystifyingrolesllmlayers, ashuach-etal-2026-masked} suggests partial functional specialisation across Transformer layers. Lower-to-middle feed-forward layers often support factual associations \cite{geva-etal-2021-transformer}, while upper layers are important for some multi-step and arithmetic computations \cite{stolfo2023a, hou-etal-2023-towards}.
These findings do not imply a strict separation between knowledge and reasoning. Instead, they motivate the hypothesis that tasks with different balances of retrieval and intermediate computation may produce different internal patterns. Motivated by this evidence, we test for systematic associations between these patterns and task demands without making causal claims.

In this paper, we distinguish knowledge retrieval from reasoning by the amount of intermediate transformation required during inference. Knowledge retrieval refers to identifying or recalling task-relevant factual or conceptual information, with little or no transformation of that information. Reasoning refers to deriving task-relevant information by composing or transforming available information through intermediate operations such as inference, calculation, comparison, or state tracking. As most tasks require both processes, we therefore use ``knowledge-centric'' and ``reasoning-centric'' to describe dominant tendencies along a continuum rather than mutually exclusive task categories.
This distinction allows us to examine whether similar task-level performance can arise from different underlying processing demands and failure modes.
Our investigation is structured around two core research questions:
\begin{itemize}
    \item RQ1: Do tasks with different balances of knowledge retrieval and reasoning produce systematically different layer-wise and decoding-time patterns in an unmodified LLM?
    \item RQ2: How do model family, parameter scale, and inference-time prompting affect performance across the knowledge-reasoning continuum?
\end{itemize}
Our contribution is not a new activation or probing metric in isolation. Instead, we introduce a task-level diagnostic framework that combines existing internal signals within a unified knowledge-reasoning analysis. Specifically, we: (i) introduce a task-centric framework, which combines static layer-wise activation statistics, decoding-time trajectories, and representation-based measures; (ii) use these signals to characterise benchmark demands and construct a knowledge-reasoning capability landscape across 16 models and 24 benchmarks; and (iii) connect the capability landscape to behavioural stress tests that reveal how prompting affects performance and robustness to shortcut-inducing perturbations.

\section{Related Work}

Our research is situated at the intersection of two critical areas: the effort to disentangle core abilities like knowledge and reasoning, and the field of mechanistic interpretability, which seeks to explain model behaviour through its internal structure.

\subsection{Behavioural Approaches}

A significant body of research has sought to separate knowledge and reasoning by analysing a model's inputs and outputs. The most prominent example is Chain-of-Thought (CoT) prompting \cite{wei2022chain}, which encourages models to externalise their reasoning steps. Subsequent frameworks have built on this by evaluating the logical quality and factual correctness of the generated reasoning chain \cite{wu2025knowledge}, often revealing that models can produce specious or unfaithful explanations \cite{press-etal-2023-measuring, turpin2023language, lanham2023measuringfaithfulnesschainofthoughtreasoning, chen2025reasoningmodelsdontsay}. While these methods provide valuable insight into a model's problem-solving process, they are fundamentally analyses of the final text output, not the internal computations that produced it.
Another line of work actively alters the model's inference process to force a separation. Some approaches fine-tune models with special tokens (e.g., \texttt{<memory>} and \texttt{<reason>}) to explicitly direct the model's processing path \cite{jin-etal-2025-disentangling-memory}. Others \cite{zhang2025system, yang2025decoupling, jia2025fast} draw inspiration from dual-process theory \cite{groves1970habituation}, prompting models to adopt distinct \textit{fast} (System 1) and \textit{slow} (System 2) thinking modes \cite{kahneman2011thinking} to disentangle intuitive knowledge retrieval from deliberate reasoning. These are powerful interventionist techniques, but they do not characterise the intrinsic behaviour of an unmodified LLM. Our work differs by aiming to diagnose the model's unaltered inference process.

\subsection{Mechanistic Foundations}

A distinct and complementary field, mechanistic interpretability, suggests that the conceptual separation between knowledge and reasoning is mirrored by a structural separation within the Transformer architecture itself \cite{zhang2024investigating, shi2024understanding, yang-etal-2025-large, gromov2025the}. This provides the theoretical foundation for our empirical approach. \citet{geva-etal-2021-transformer} demonstrated that Feed-Forward Networks (FFNs) function as key-value memories, storing factual associations learned during training, particularly in the lower-to-mid layers of the model. This finding has been extended by research that successfully edits specific factual knowledge within the FFN weights of mid-layer transformers \cite{meng2022locating, meng2023massediting}, providing a mechanistic basis for where parametric knowledge resides.
In contrast, causal analysis and probing studies have identified the upper layers of the network as being critical for multi-step, compositional, and arithmetic reasoning \cite{stolfo2023a, hou-etal-2023-towards}. For instance, some studies interpret LLMs by focusing on how the information is processed between layers \cite{olsson2022context, nanda2023progress}. Related work further suggests that information encoded in internal representations may not always be reflected in observable model behaviour \cite{liu2026broken}. These higher-level processes are understood to integrate and manipulate information that has been retrieved or processed by the lower layers.
These findings motivate a testable hypothesis rather than a strict localisation claim: tasks that rely more on direct retrieval and tasks that require more intermediate computation may produce different patterns across layers. Because representations can become less distinguishable in deeper causal Transformers \cite{hou-etal-2023-towards}, final-layer states may obscure differences that appear earlier. We therefore analyse the full layer-wise evolution of internal states to identify where task-related differences emerge and whether they persist.

\subsection{Towards Holistic Evaluation}

Behavioural methods analyse the model from the outside, while mechanistic work ``reverse engineer'' the function of internal components in isolation \cite{hou-etal-2023-towards, zhao2024how, zhu2025surveylatentreasoning, zhao2025towards, wang2025adversarial, hong2026beyond}. Our work bridges this divide by operationalising the findings from mechanistic interpretability.
Instead of focusing on model-centric properties, such as a model's self-awareness of its knowledge gaps \cite{yin-etal-2023-large, manakul-etal-2023-selfcheckgpt} or the procedural limits of its CoT capabilities \cite{dziri2023faith, chen2024unlocking}, our approach is task-centric.
We use internal activation patterns to characterise the relative retrieval and reasoning demands that a task elicits in an LLM. We then compare models according to how their internal states and behavioural performance respond to these demands. This provides a unified view of model performance across the knowledge-reasoning continuum.
Unlike prior studies that use individual interpretability techniques to localise specific behaviours, we combine complementary internal signals to compare task demands across benchmarks and model families. Rather than introducing a new metric, its novelty lies in using these signals together to capture different aspects of task-level processing.

\section{Method and Experimental Setup}
\label{sec:experiment_setting}

\subsection{Benchmarks and Taxonomy}
\label{sec:benchmark_method}

We evaluate 24 benchmarks covering factual and biomedical QA, scientific knowledge, commonsense and logical reasoning, reading comprehension, symbolic reasoning, and mathematical problem solving. Appendix~\ref{sec:datasets} reports the complete benchmark list, evaluation formats, numbers of examples, and citations.
We classify each benchmark by its dominant task demand. A benchmark is \textit{knowledge-centric} when most items primarily require factual retrieval or verification with limited intermediate transformation. A benchmark is \textit{reasoning-centric} when most items require composition, inference, calculation, comparison, or state tracking. This assignment is based on the benchmark design and required operations and is made independently of the analyses conducted in this paper. The labels describe dominant tendencies rather than disjoint categories, as many benchmarks require both knowledge and reasoning. Appendix~\ref{sec:benchmark_taxonomy} provides the complete assignment and discusses mixed cases.

\paragraph{Scope.}
The capability landscape includes both multiple-choice and free-form benchmarks, but probing, representation similarity, and \textsc{Nota} (\textit{none of the above}) analyses use multiple-choice tasks because fixed candidate sets enable controlled representation extraction and answer-option perturbation. We restrict the study to English to avoid confounding task demands with differences in tokenisation, language coverage, and translation. These choices provide a controlled diagnostic setting.

\subsection{Profiling Representations and Internal States of Models}
\label{sec:prism}

We propose a task-centric framework for characterising the model-internal processing elicited by a task in an unmodified LLM. It does not directly measure knowledge or reasoning capability; rather, it compares internal patterns across tasks with different balances of retrieval and intermediate transformation. We combine layer-wise attention entropy and feed-forward activation sparsity, their trajectories during decoding, and representation-based evidence from probing and correct–incorrect answer comparisons.

\paragraph{Static Profile.}
We define a static profile as the benchmark-level pattern of layer-wise attention entropy and activation sparsity measured before answer generation.
For each benchmark, we process all available questions with Qwen2.5-7B-Instruct without generating an answer. We select Qwen2.5-7B-Instruct as the reference model because it provides a tractable mid-sized, open-weight setting for extracting layer-wise internal states across all 24 benchmarks. To reduce the risk that the observed representational patterns are specific to this model family, the subsequent probing and representation analyses additionally examine Llama3.1-8B-Instruct and OLMo-3-7B-Instruct. We extract the layer-output hidden state and attention distribution at the final input-token position.
No prompt template or instruction is added in this analysis, so all 24 benchmarks share the same input format.
We use attention entropy to characterise how concentrated or diffuse the model's attention is across the input at each layer. The layer-wise entropy profile therefore serves as a descriptive measure of how selectively the model allocates attention when processing tasks with different retrieval and reasoning demands.

Let \(a^{(l,h)}_{i,j}\) denote the attention weight assigned to input position \(j\) by head \(h\) in layer \(l\), when the query is the final input token of example \(i\). We first average the attention distribution across the \(M\) heads:
\begin{equation}
\bar{a}^{(l)}_{i,j}
=
\frac{1}{M}
\sum_{h=1}^{M}
a^{(l,h)}_{i,j}.
\label{eq:head_averaged_attention}
\end{equation}

We normalise this distribution as
\begin{equation}
p^{(l)}_{i,j}
=
\frac{\bar{a}^{(l)}_{i,j}}
{\sum_k \bar{a}^{(l)}_{i,k}+\delta},
\label{eq:normalised_attention}
\end{equation}
where \(\delta=10^{-10}\). The layer-wise attention entropy is then
\begin{equation}
H_l
=
-\frac{1}{N}
\sum_{i=1}^{N}
\sum_j
p^{(l)}_{i,j}
\log
\left(
p^{(l)}_{i,j}+\delta
\right),
\label{eq:attention_entropy}
\end{equation}
where \(N\) is the number of examples.

We use activation sparsity to characterise how selectively the model's hidden representation is activated at each layer. Higher sparsity indicates that a larger proportion of hidden dimensions remain near zero, providing a descriptive measure of how concentrated or distributed the layer-wise representation is across tasks.

Let \(u^{(l)}_{i,r}\) denote dimension \(r\) of the layer-output
hidden state at the final input-token position, and let \(d\) denote
the model hidden size. We compute hidden state sparsity as
\begin{equation}
S_l
=
\frac{1}{N}
\sum_{i=1}^{N}
\frac{1}{d}
\sum_{r=1}^{d}
\mathbb{I}
\left(
\left|u^{(l)}_{i,r}\right| < 0.01
\right).
\label{eq:hidden_state_sparsity}
\end{equation}

The dataset-level static feature vector is
\begin{equation}
F_d
=
\left[
H_1,\ldots,H_L,
S_1,\ldots,S_L
\right],
\label{eq:static_footprint}
\end{equation}
where \(L\) is the number of Transformer layers. These features
summarise the final-token attention distribution and layer-output
hidden state. They are descriptive and are not interpreted as causal
measures of knowledge retrieval or reasoning.

\paragraph{Decoding-Time Trajectories.}
This analysis is intended to characterise how internal processing evolves during answer generation, rather than only how it differs across layers at input time.
We trace the same metrics over generated tokens for all available SciQ examples using Qwen2.5-7B-Instruct, and repeat the analysis on two reasoning-centric benchmarks, MathQA and LogiQA.
We use SciQ as the primary controlled diagnostic dataset because its questions are short and relatively homogeneous, reducing confounds from variable generation lengths and heterogeneous reasoning strategies. The two reasoning-centric benchmarks test whether the resulting signal extends beyond knowledge retrieval.
All three benchmarks use the retrieve-then-answer prompt given in Figure~\ref{fig:trajectory_prompt}. We compare the token-wise mean trajectories and within-trajectory variability of attention entropy and activation sparsity between correct and incorrect generations.
We report layers 3, 14, and 22 as representative lower, middle, and upper layers. Generation uses temperature \(0.7\), top-\(p\) \(0.8\), and a maximum of 1,024 new tokens; generations that are truncated or contain no parsable answer are excluded. Signals are read at the positions of the generated tokens: after sampling, the prompt and sampled tokens are passed through the model once, which, under causal masking, reproduces the internal state at each decoding step.
To compare responses of different lengths, we sample $P=101$ uniformly spaced positions from each generation of length \(G\). For \(G>1\), the selected
positions are
\begin{equation}
i_k
=
\left\lfloor
\frac{k(G-1)}{P-1}
\right\rfloor
\label{eq:trajectory_subsampling}
\end{equation}
where $k\in\{0,\ldots,P-1\}$.

\subsection{Probing and Representation Analysis}
\label{sec:representation_method}

To complement the static profile and decoding-time trajectory analyses, we use linear probing and representation similarity analysis to examine two questions: (1) at which layers is the correct answer option linearly recoverable from pre-generation hidden states, and (2) how distinctly does the model represent a question paired with its correct answer versus the same question paired with an incorrect distractor?

We conduct these probing and representation analyses on a balanced diagnostic subset of four multiple-choice benchmarks: two knowledge-centric tasks (BioMixQA and SciQ) and two reasoning-centric tasks (LogiQA and MathQA). We use multiple-choice benchmarks because their fixed candidate sets enable controlled extraction and comparison of correct and incorrect answer representations. The analyses are conducted using Llama3.1-8B-Instruct, Qwen2.5-7B-Instruct, and OLMo-3-7B-Instruct.

\paragraph{Layer-Wise Probing.}
For each question, we perform one forward pass using the templates in Figure~\ref{fig:probing_prompt_mcp} and extract the hidden state at the final input position before answer generation. At each layer, we train a logistic regression classifier to predict the correct answer option using 5-fold cross-validation and report average held-out accuracy.
We compare this accuracy with the majority-class baseline, obtained by always predicting the most frequent correct answer label in each benchmark.
We track accuracy across layers to examine when the correct answer becomes linearly recoverable in each task group.
Accuracy close to this baseline does not show that reasoning information is absent. It indicates only that the selected linear probe does not reliably recover the answer under the given prompt and token position.
The task groups use different templates, so task type and template choice are confounded. We therefore run every benchmark under both templates to compare the task groups under the same prompting conditions. The model, questions, probing procedure, and token position are held fixed. From these forward passes, we record the head-averaged attention of the final input token (Eq.~\ref{eq:head_averaged_attention}). We report the share assigned to the question and answer options, whose text is identical in both templates.

\paragraph{Representation Similarity.}
For each question \(q\), we append either the correct answer \(a^{+}\) or an incorrect distractor \(a^{-}\):
\begin{equation}
x^{+}=[q;a^{+}],
\qquad
x^{-}=[q;a^{-}].
\end{equation}

We extract their final-token hidden states at each layer and compute an adjusted cosine similarity:
\begin{equation}
\operatorname{sim}_{\mathrm{adj}}(h^{+},h^{-})
=
\cos(h^{+},h^{-})
-
\mathbb{E}_{i,j}
\left[
\cos(h_i,h_j)
\right],
\end{equation}
where the expectation is computed over randomly paired completions from the same benchmark.
We use this measure to characterise how distinctly the model represents the correct answer relative to a plausible incorrect alternative. Comparing this separation across task types allows us to test whether knowledge-centric and reasoning-centric benchmarks differ systematically in the geometry of their answer representations.
This measure captures relative representational similarity and does not establish causal separation between correct and incorrect answers.

\subsection{Capability Scores and Evaluation}
\label{sec:score_construction}

Let \(s_{m,p,d}\) denote the performance of model \(m\), under prompting condition \(p\), on benchmark \(d\). We compute the knowledge and reasoning scores as unweighted averages over these reported evaluation settings:
\begin{align}
K_{m,p}
&=
\frac{1}{|\mathcal{D}_{K}|}
\sum_{d \in \mathcal{D}_{K}}
s_{m,p,d},
\label{eq:knowledge_score}
\\
R_{m,p}
&=
\frac{1}{|\mathcal{D}_{R}|}
\sum_{d \in \mathcal{D}_{R}}
s_{m,p,d}.
\label{eq:reasoning_score}
\end{align}
where ($|\mathcal{D}_K|=10$) and ($|\mathcal{D}_R|=20$). Each reported setting therefore contributes equally to the corresponding capability score, irrespective of the number of examples it contains.

For either dimension \(X \in \{K,R\}\), let
\begin{align}
X_{\min} &= \min_{m,p} X_{m,p}, \\
X_{\max} &= \max_{m,p} X_{m,p}.
\end{align}
\label{eq:min_max}
and apply min-max normalisation:
\begin{equation}
\widetilde{X}_{m,p}
=
\frac{X_{m,p}-X_{\min}}
     {X_{\max}-X_{\min}}.
\label{eq:score_normalisation}
\end{equation}
where min/max are taken jointly over model $\times$ prompting condition (Standard and CoT), separately for each dimension.
Evaluation scores are expressed on a common $[0,1]$ scale. We use accuracy for classification and multiple-choice settings and exact match for the free-form mathematical benchmarks GSM8K, AIME, MATH500, and AMC.

We evaluate open-weight, instruction-tuned models from the Qwen, Llama, Gemma, OLMo, GLM, and GPT-OSS families. The models range from 2B to more than 30B parameters and include both dense and Mixture-of-Experts architectures. Full model details are provided in Appendix~\ref{sec:models}.

All evaluations use EvalScope~\cite{evalscope_2024}, with SGLang~\cite{zheng2024sglang} as the inference backend. We follow the benchmark-specific zero-shot and few-shot settings provided by the evaluation framework. We use accuracy for classification and multiple-choice tasks and exact match for free-form mathematical tasks. Standard and CoT conditions use the same evaluation configuration except for the prompting instruction.

\begin{figure*}[t]
    \centering
    \includegraphics[width=1.0\linewidth]{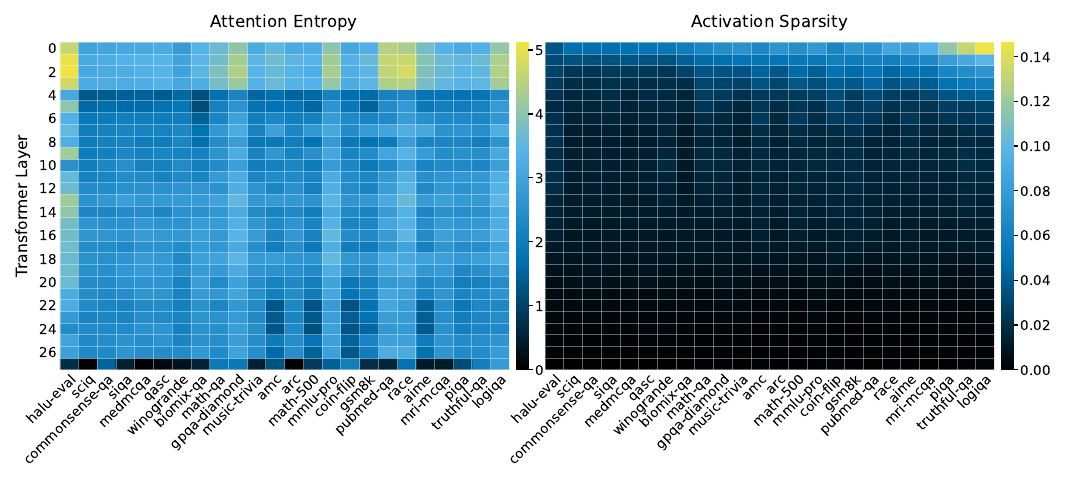}
    \caption{Layer-wise static internal profiles of Qwen2.5-7B-Instruct across 24 benchmarks.
    The heatmaps show attention entropy (left), where higher values indicate more diffuse attention, and activation sparsity (right), defined as the proportion of near-zero feed-forward activations.
    Benchmarks are ordered by the first principal component of the activation sparsity matrix to expose cross-benchmark variation in their layer-wise profiles.
    }
    \label{fig:neuron}
\end{figure*}

\begin{figure*}[t]
    \centering
    \includegraphics[width=1\linewidth]{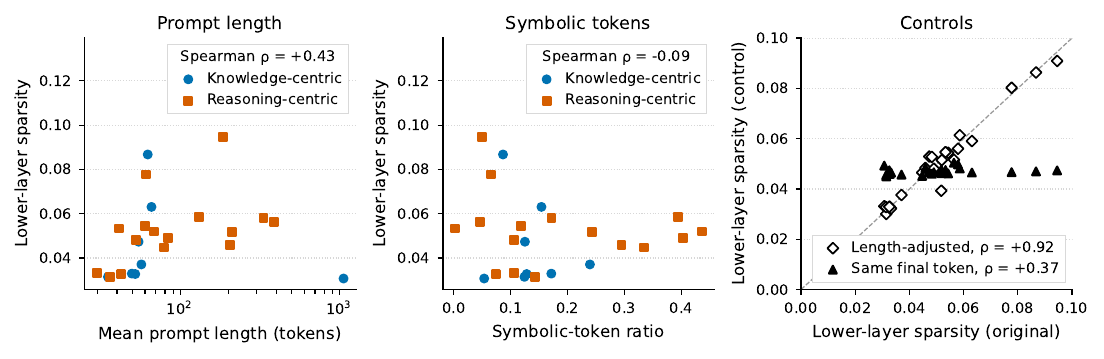}
    \caption{Controls for the lower-layer activation sparsity in Figure~\ref{fig:neuron} (Qwen2.5-7B-Instruct; mean over layers 0-3 at the final input token). Left and middle: benchmark means against mean prompt length and against the share of prompt tokens that contain a digit or mathematical symbol. Right: benchmark means under two controls against the original values. Length-adjusted values are predicted at a fixed length from a within-benchmark regression on log length; same-final-token values append ``\texttt{\textbackslash{}nAnswer:}'' to every prompt. The dashed line marks equality.}
    \label{fig:sparsity_controls}
\end{figure*}

\begin{figure*}[t]
    \centering
    \includegraphics[width=0.9\linewidth]{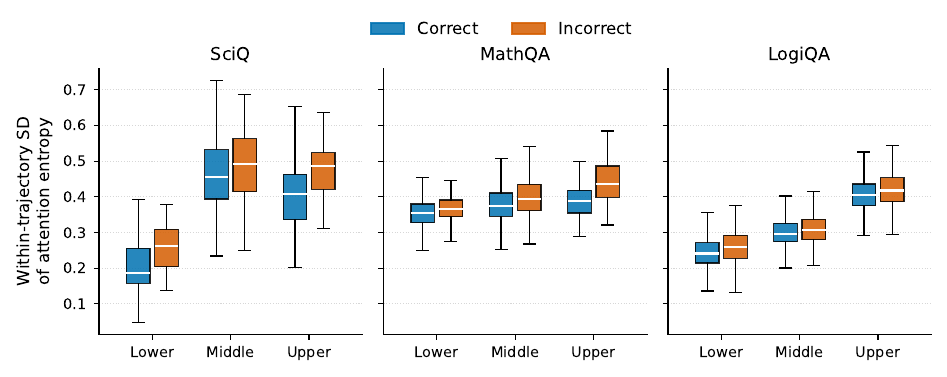}
    \caption{Within-trajectory variability of attention entropy for Qwen2.5-7B-Instruct on SciQ, MathQA, and LogiQA. For each generation, the standard deviation of attention entropy is computed over the 101 aligned generated-token positions at the lower, middle, and upper depths (layers 3, 14, and 22).}
    \label{fig:trajectory_reasoning}
\end{figure*}

\begin{figure*}[t]
    \centering
    \includegraphics[width=1\linewidth]{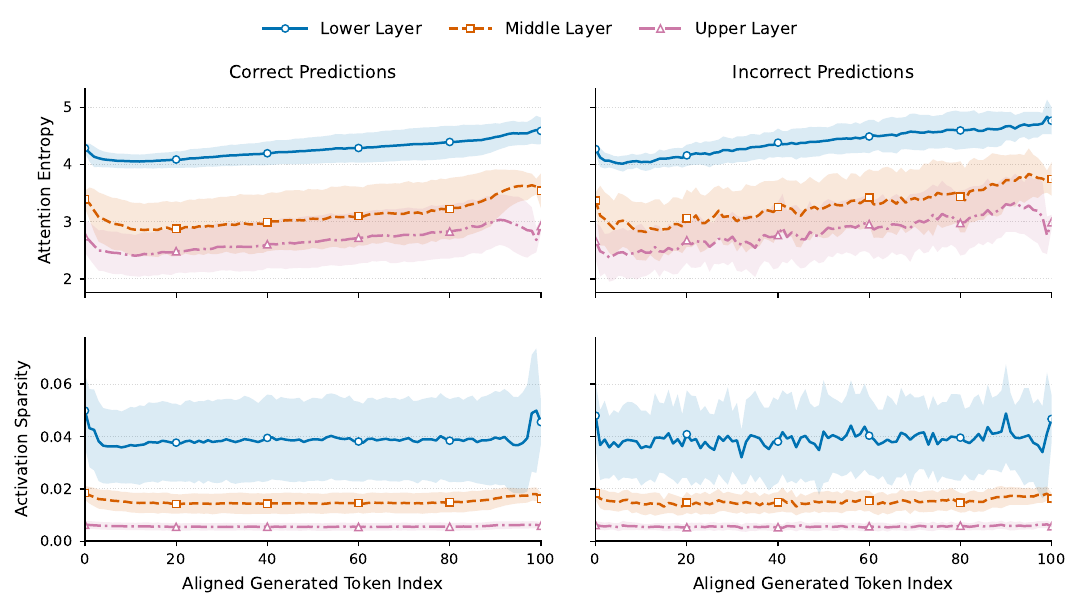}
    \caption{Decoding-time trajectory analysis on SciQ using Qwen2.5-7B-Instruct. Rows show attention entropy (top) and activation sparsity (bottom). Columns separate correct (left) and incorrect (right) predictions. Curves show token-wise mean trajectories at representative lower, middle, and upper layers (layers 3, 14, and 22). Shaded regions show $\pm 1$ standard deviation across examples within each correctness group. Generated sequences are length-normalised to 101 aligned token positions by uniform subsampling. Incorrect generations are associated with greater within-trajectory variability in attention entropy, particularly at the lower and upper layers. The one-sided tests do not detect greater within-trajectory variability in activation sparsity for incorrect generations (Table~\ref{tab:trajectory_variability}).}
    \label{fig:trajectory}
\end{figure*}

\begin{figure*}[t]
    \centering
    \includegraphics[width=1\linewidth]{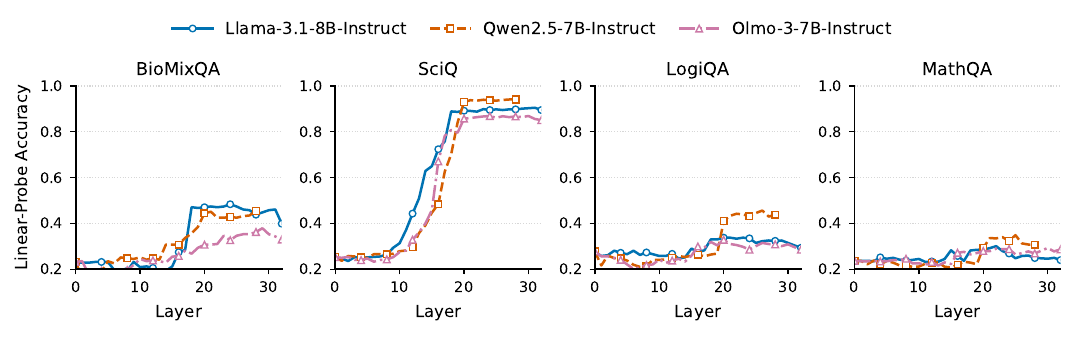}
    \caption{Layer-wise linear probing accuracy under the \emph{multiple-choice} setting across three model families.
    Knowledge-centric datasets show increasing probing accuracy in the mid-to-late layers, indicating that the correct option becomes more linearly recoverable.
    Prompts are shown in Figure~\ref{fig:probing_prompt_mcp}.}
    \label{fig:mcp}
\end{figure*}

\begin{figure*}[t]
    \centering
    \includegraphics[width=1\linewidth]{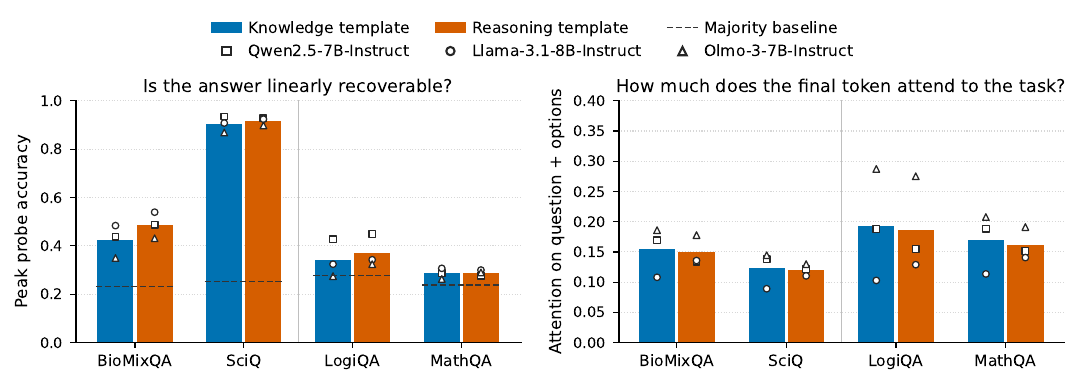}
    \caption{Template-swap control for the probing analysis. Each benchmark is run under both the knowledge-centric and the reasoning-centric template of Figure~\ref{fig:probing_prompt_mcp}. Bars show the mean over Llama3.1-8B-Instruct, Qwen2.5-7B-Instruct, and OLMo-3-7B-Instruct; markers show individual models. Left: peak layer-wise probing accuracy; dashed lines mark the majority-class baseline. Right: share of the final token's attention, averaged over heads and layers, on the question and answer options.}
    \label{fig:template_swap}
\end{figure*}

\begin{figure*}[t]
    \centering
    \includegraphics[width=1\linewidth]{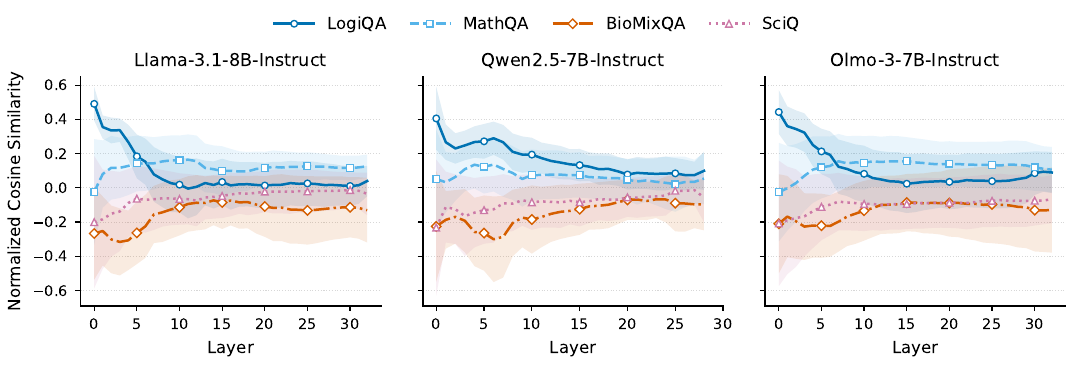}
    \caption{
    Representation geometry of correct and distractor answers.
    Curves show the adjusted cosine similarity between hidden states for prompts appended with the ground-truth answer versus an incorrect distractor; shaded regions indicate variability across examples. Lower values indicate greater representational separation. Knowledge-centric benchmarks (BioMixQA and SciQ) generally show lower similarity than reasoning-centric benchmarks (LogiQA and MathQA) across the three models.
    }
    \label{fig:similarity}
\end{figure*}

\begin{figure*}[tp]
    \centering
    \includegraphics[width=1.0\linewidth]{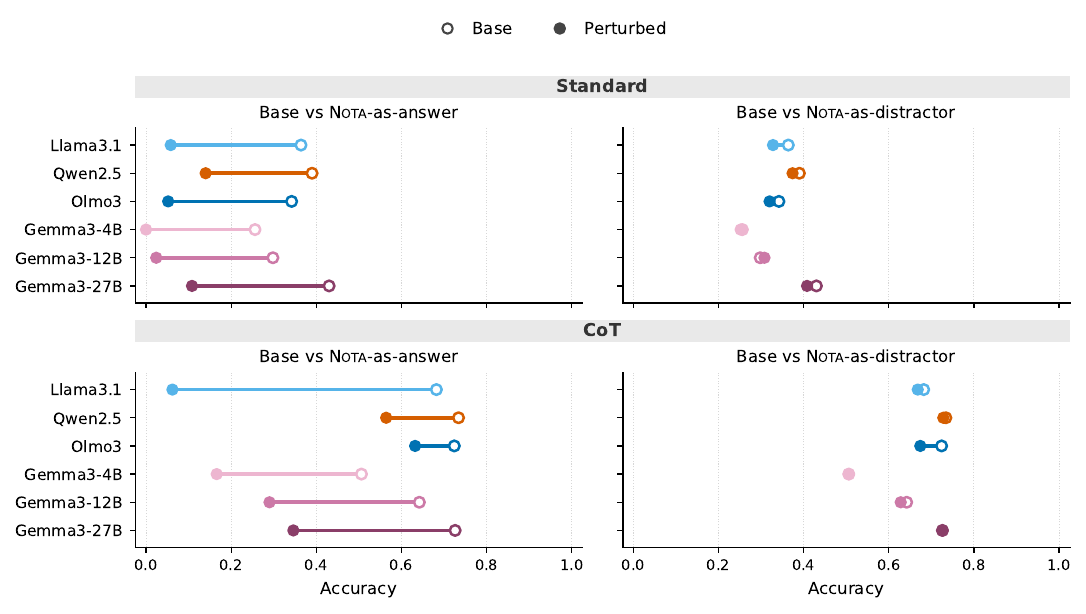}
    \caption{
    Robustness to \textsc{Nota} perturbations on MathQA under Standard and CoT prompting. Hollow markers ($\circ$) indicate Base accuracy with the original answer options, while solid markers ($\bullet$) show accuracy after the \textsc{Nota}-as-answer or \textsc{Nota}-as-distractor perturbation; connecting lines indicate the resulting performance change. \textsc{Nota}-as-answer produces substantially larger degradation than \textsc{Nota}-as-distractor for most models, with the effect varying across prompting conditions.
    }
    \label{fig:mathqa_dumbbell}
\end{figure*}

\begin{figure*}[tp]
    \centering
    \includegraphics[width=1.0\linewidth]{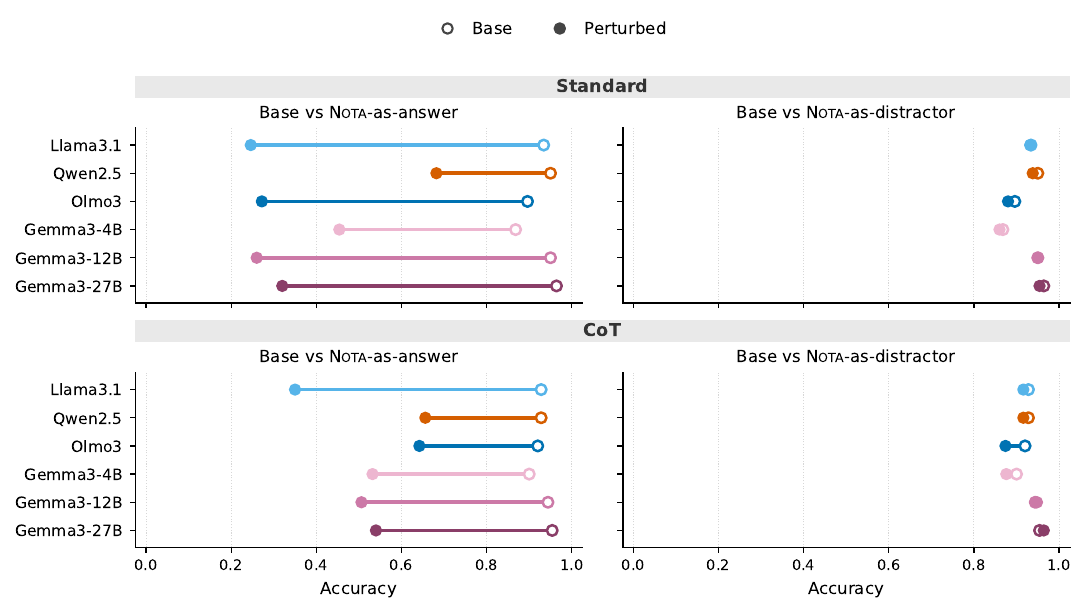}
    \caption{
    Robustness to \textsc{Nota} perturbations on SciQ under Standard and CoT prompting. Hollow markers ($\circ$) indicate Base accuracy with the original answer options, while solid markers ($\bullet$) show accuracy under the \textsc{Nota}-as-answer or \textsc{Nota}-as-distractor condition; connecting lines indicate the performance change. \textsc{Nota}-as-answer causes a pronounced accuracy drop across models, whereas adding \textsc{Nota} as a distractor has comparatively little effect.
    }
    \label{fig:sciq_dumbbell}
\end{figure*}

\begin{figure*}[t]
    \centering
    \includegraphics[width=0.95\linewidth]{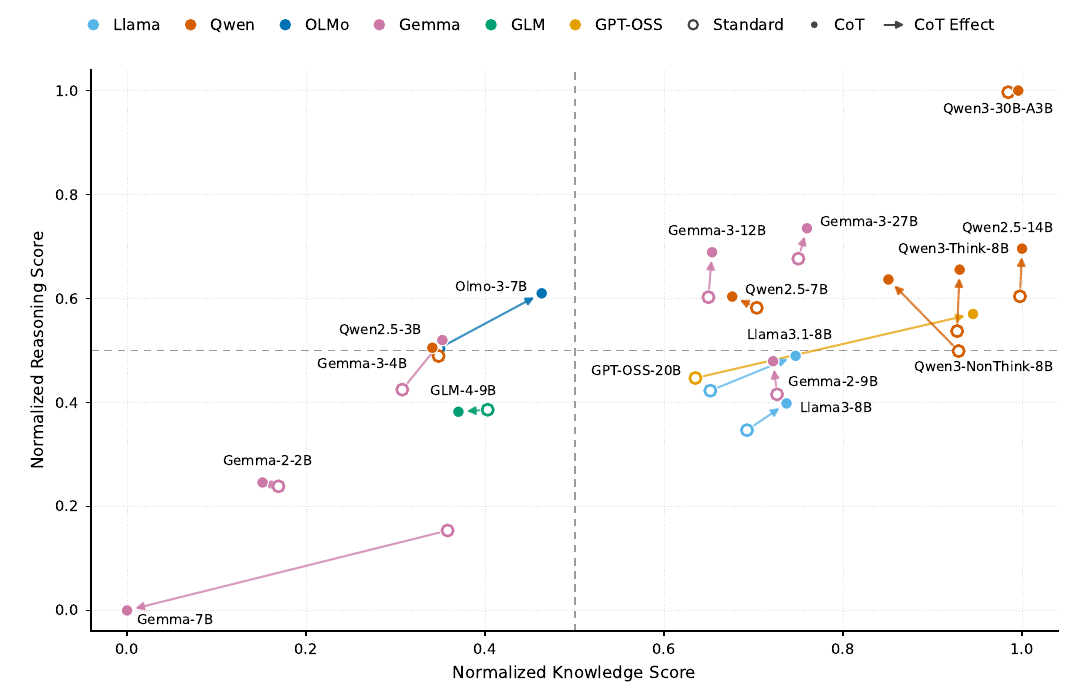}
    \caption{
    For each model and prompting condition, raw performance is first averaged over the corresponding knowledge or reasoning evaluation settings (Eqs.~\ref{eq:knowledge_score}-\ref{eq:reasoning_score}). Each resulting dimension is then min-max normalised jointly across all model and prompting-condition pairs (Eq.~\ref{eq:score_normalisation}). Hollow ($\circ$) and solid ($\bullet$) markers denote Standard and CoT prompting, respectively, and arrows indicate the change induced by CoT.
    }
    \label{fig:landscape}
\end{figure*}

\section{Knowledge-Reasoning Patterns}
\label{sec:knowledge_reasoning_pattern}

\paragraph{RQ1: Do tasks with different balances of knowledge retrieval and reasoning produce systematically different layer-wise and decoding-time patterns in an unmodified LLM?}

\paragraph{Internal Mechanics.}
We first examine whether an LLM exhibits a consistent, data-driven internal pattern that differentiates tasks. To establish a baseline internal profile, we perform a \textit{question-only} analysis using Qwen2.5-7B-Instruct. For each dataset, we feed the model only the input question (without generating an answer) and summarise its layer-wise internal activity into a single static profile. This design isolates task comprehension and retrieval demands from decoding-dependent effects such as variable generation length, sampling stochasticity, or branching during answer construction. As a result, differences in the profile reflect how the model processes a fixed input, rather than decoding-dependent variability.

We compute attention entropy and activation sparsity at each layer. Attention entropy measures how concentrated the attention distribution is. Higher values indicate more diffuse attention. Activation sparsity is the proportion of feed-forward activations that are close to zero. We concatenate these values across layers to form a feature vector for each dataset.
Figure~\ref{fig:neuron} shows these profiles for all 24 datasets. We order the datasets by the first principal component of the activation sparsity matrix. This produces a smooth progression in sparsity across datasets and layers. Under this ordering, activation sparsity shows a clearer pattern in the lower layers than attention entropy.
LogiQA, TruthfulQA, and PIQA have the highest sparsity in the lower layers. The mathematical benchmarks AIME, GSM8K, MATH500, AMC, and MathQA fall in the middle of the range. The lowest values occur in both task groups. These include knowledge-centric benchmarks such as SciQ and MedMCQA and reasoning-centric benchmarks such as CommonsenseQA and SIQA. Figure~\ref{fig:neuron} describes how the model processes the input of each benchmark. The controls below show that adding the same suffix to every input reduces the differences in lower-layer sparsity across benchmarks. This indicates that these measurements are sensitive to how the input ends. We therefore use probing, decoding trajectories, representation geometry, and behavioural analyses to compare knowledge and reasoning.

\paragraph{Prompt Length and Token-Composition Controls.}
Because benchmarks differ in input length, vocabulary, and formatting, we test whether these input properties explain the differences in lower-layer sparsity in Figure~\ref{fig:neuron}. We run the model again on the same questions from each benchmark, one question at a time. We record the prompt length and token composition for each question (Figure~\ref{fig:sparsity_controls}).
Across the 24 benchmarks, lower-layer sparsity (layers 0 to 3) has a moderate correlation with mean prompt length (Spearman $\rho=0.43$). It shows little correlation with the proportion of tokens containing digits or mathematical symbols ($\rho=-0.09$). We fit a regression of sparsity on log prompt length within each benchmark. We then estimate sparsity at a common prompt length, which is the median across all questions. This adjustment leaves the benchmark ordering nearly unchanged ($\rho=0.92$ with the original ordering).
We measure sparsity at the final input token. To test whether this token affects the measurements, we append the same suffix (``\texttt{\textbackslash{}nAnswer:}'') to every prompt. This makes all prompts end with the same token. The standard deviation of lower-layer sparsity across benchmarks decreases. The rank correlation with the original benchmark ordering falls to $\rho=0.37$.
We find no statistically significant difference in lower-layer sparsity between knowledge-centric and reasoning-centric benchmarks. This holds both before and after the suffix is added (Mann-Whitney $p=.19$ and $p=.21$, respectively). These results indicate that the lower-layer differences in Figure~\ref{fig:neuron} depend on how each benchmark's input ends. We therefore do not draw conclusions about knowledge and reasoning from lower-layer sparsity alone.

\paragraph{Decoding-time Trajectories.}
Using the controlled SciQ setup described in \S\ref{sec:prism}, we examine how attention entropy and activation sparsity evolve during generation for correct and incorrect predictions. SciQ is additionally evaluated with a structured retrieve-then-answer prompt (Figure~\ref{fig:trajectory_prompt}), providing a consistent temporal scaffold for interpreting these trajectories.

Figure~\ref{fig:trajectory} visualises the evolution of internal signals during generation on SciQ. Here, the \textit{mean trajectory} represents the token-wise average of the metric across examples after length normalisation. A key observation is that \textit{mean levels alone} often fail to distinguish correct from incorrect outputs. The diagnostic signal lies primarily in differences in \textit{within-trajectory variability}. Correct predictions display a clearer late-stage drop in attention entropy (reflecting answer consolidation). However, the separation in mean levels is modest. Incorrect generations show greater within-trajectory variability in attention entropy.
On SciQ, incorrect generations are associated with greater within-trajectory variability in attention entropy, particularly at the lower and upper layers (Table~\ref{tab:trajectory_variability}). The one-sided tests do not detect greater within-trajectory variability in activation sparsity for incorrect generations than for correct ones.

Figure~\ref{fig:trajectory_reasoning} shows that this pattern is not specific to SciQ. As on SciQ, incorrect generations on MathQA and LogiQA show a larger within-trajectory standard deviation of attention entropy than correct ones. Under the one-sided Mann-Whitney test used in Table~\ref{tab:trajectory_variability}, the difference is significant at the lower and upper layers on all three benchmarks ($q<.01$).
The difference is also significant at the middle layer on MathQA and LogiQA. Activation sparsity provides no consistent signal across benchmarks. Its within-trajectory variability differs between correct and incorrect generations at some layers on MathQA and LogiQA. On SciQ, the one-sided tests do not detect greater sparsity variability in incorrect generations.

\paragraph{Probing and Geometric Analysis.} The trajectory analysis suggests that correct and incorrect generations differ more in within-trajectory variability than in mean signal levels, particularly for attention entropy. We next examine how reliably a linear probe can recover the correct answer option from a \textit{static} hidden state before generation in a multiple-choice setting.

Figure~\ref{fig:mcp} reveals a clear contrast in pre-generation answer recoverability across the two task groups. On the knowledge-centric benchmarks BioMixQA and SciQ, probing accuracy rises consistently through the mid-to-late layers across the three model families, indicating that the correct option becomes progressively more linearly accessible before answer generation. In contrast, probing accuracy on LogiQA and MathQA stays close to the majority-class baseline for Llama3.1-8B-Instruct and OLMo-3-7B-Instruct (peak 0.29-0.34 against baselines of 0.24 and 0.28). Qwen2.5-7B-Instruct is the exception: its accuracy rises above chance from around layer 20, peaking at 0.46 on LogiQA and 0.35 on MathQA, although its gains over the baseline (0.18 and 0.11) remain smaller than on BioMixQA and SciQ (0.22 and 0.69).

Under the probing protocol defined in \S\ref{sec:representation_method}, these results indicate that answer-discriminative information is more linearly recoverable from pre-generation hidden states for the two knowledge-centric tasks than for the two reasoning-centric tasks. This should not, however, be interpreted as evidence that reasoning information is absent from the latter representations; rather, the correct answer is not reliably recoverable from them using the selected prompt, final-token representation, and linear probe.

\paragraph{Effect of Prompt Structure.}
Figure~\ref{fig:template_swap} shows that using different templates for the two task groups does not fully explain the probing contrast in Figure~\ref{fig:mcp}. Averaged over Llama3.1-8B-Instruct, Qwen2.5-7B-Instruct, and OLMo-3-7B-Instruct, BioMixQA and SciQ retain higher peak probing accuracy than LogiQA and MathQA under both templates. The reasoning-centric template attracts more attention to its instructions and response cues. Attention to the question and answer options remains nearly unchanged. The extra instructions therefore change how the model allocates attention without closing the gap in answer recoverability. This supports the knowledge-reasoning comparison for these benchmarks under the two tested templates.

\paragraph{Distinctness of Representations.} While linear probing reveals whether correctness is \textit{separable} from hidden states, it does not quantify how \textit{far apart} correct and incorrect completions lie in representation space.
We therefore compare the adjusted correct–incorrect representation similarity defined in \S\ref{sec:representation_method} across the four diagnostic benchmarks.
In Figure~\ref{fig:similarity}, BioMixQA and SciQ show lower adjusted cosine similarity, whereas LogiQA and MathQA show higher similarity. This pattern is consistent with greater semantic separation between factual answer options and greater lexical or structural overlap between reasoning options. It may reflect both task demands and distractor construction. The result therefore shows an association between benchmark type and representation geometry, not a causal or structural separation between truth and falsehood.
One possible explanation is that factual distractors often differ from correct answers in entity or semantic content, whereas reasoning distractors are frequently near-miss solutions with similar wording and structure. The observed geometry may therefore reflect both task demands and answer-option construction. It should not be interpreted as evidence that truth and falsehood are causally or structurally separated in the early layers.

The analyses in this section provide complementary forms of evidence. The static analysis compares all benchmarks at input time, the trajectory analysis examines correct and incorrect generations on SciQ, MathQA, and LogiQA, and the probing and geometric analyses examine four multiple-choice benchmarks. Their agreement supports an association between benchmark demands and internal processing patterns. However, prompt format, answer semantics, distractor construction, and dataset composition may also contribute to the observed differences. Because we do not perform activation patching, layer ablation, or causal tracing, these results identify diagnostic associations rather than causal mechanisms.

\section{Knowledge-Reasoning Landscape}
\label{sec:capability}

We now turn from internal task characterisation to model-level capability profiling. RQ2 examines how model family, scale, and prompting shift models along the knowledge–reasoning dimensions and whether apparent gains remain robust to answer-option perturbations.

\paragraph{RQ2.1: How do the knowledge and reasoning profiles of LLMs vary with model family and scale?}

\paragraph{Model Family.}
Figure~\ref{fig:landscape} shows the patterns of different model families in the knowledge-reasoning plane: models from the same family tend to occupy similar regions, indicating that successive architectural revisions can influence performance as much as, and sometimes more than, model size. More recent families tend to achieve higher scores on both axes, placing them nearer the upper-right region of the plane. In particular, Qwen3 variants are among the strongest overall performers, while Gemma3 variants consistently outperform earlier Gemma versions along both knowledge and reasoning axes. Within the Gemma family, newer models exhibit a near-linear progression in the knowledge-reasoning plane.

\paragraph{Scaling.} Within a given family, increasing model size improves performance, but the gains are uneven across the two axes. Larger models move rightward more consistently, indicating that knowledge scores tend to increase more reliably with scale. Improvements in reasoning are less monotonic and often exhibit smaller marginal returns at larger sizes. The Gemma3 series illustrates this pattern: scaling from 3B to 27B yields a substantial gain in knowledge but only a modest improvement in reasoning. At the same time, smaller models (e.g., Gemma3-4B, Qwen2.5-3B) already achieve competitive reasoning scores.
Within the evaluated model families, knowledge scores increase more consistently with parameter count than reasoning scores. However, model size co-varies with architecture, training data, instruction tuning, and other design choices. These comparisons therefore do not isolate a causal effect of scale. The result should be interpreted as a descriptive pattern: larger models show more reliable gains on the knowledge axis, whereas reasoning gains vary more across model families.

\paragraph{RQ2.2: What is the impact of CoT prompting on the internal knowledge-reasoning balance, and does it introduce performance trade-offs?}

\paragraph{Knowledge-Reasoning Trade-off.} Figure~\ref{fig:landscape} compares each model under standard prompting (hollow marker $\circ$) and CoT prompting (solid marker $\bullet$), with an arrow ($\rightarrow$) showing the change. The main pattern is simple: \emph{most arrows point upward}, meaning CoT usually increases the reasoning score. The change in knowledge is smaller and less consistent: for many models the arrow is close to vertical (little change in knowledge), while for several models the arrow also moves to the right (a small gain in knowledge alongside improved reasoning). However, CoT is not always beneficial. A clear counterexample is Gemma-7B, which moves down-left under CoT, indicating a drop in both reasoning and knowledge.
CoT increases reasoning scores in 14 of the 16 checkpoints and knowledge scores in 9. Paired model-level tests detect a significant increase in reasoning scores but do not detect a significant change in knowledge scores (Appendix~\ref{sec:statistical_validation}, Table~\ref{tab:cot_effect}). CoT can still reduce performance for individual models, as illustrated by Gemma-7B.

\paragraph{RQ2.3: Does CoT prompting mitigate parametric shortcuts?}

\paragraph{Experiment Settings.}
To disentangle \emph{parametric shortcuts} (answering by matching memorised option patterns) from \emph{dynamic reasoning} (deriving an answer independent of the option set), we adopt a ``\textit{none of the above}'' (\textsc{Nota}) perturbation stress test, inspired by \citet{wang-etal-2025-drivel}. We evaluate two complementary perturbations under zero-shot multiple-choice evaluation: (i) \textsc{Nota}-as-answer, where we remove the ground-truth option and make \textsc{Nota} the correct choice; and (ii) \textsc{Nota}-as-distractor, where we inject \textsc{Nota} as an additional incorrect option. We compare Standard prompting against CoT prompting across six instruction-tuned LLMs (Llama3.1, Qwen2.5, Olmo3, Gemma3 4B/12B/27B) on MathQA and SciQ. We report accuracy and visualise robustness as the gap between Base (original options) and each perturbed condition (Figures~\ref{fig:mathqa_dumbbell} and \ref{fig:sciq_dumbbell}).

\paragraph{Robustness to Option Manipulation.}
On the MathQA dataset, CoT frequently improves robustness to option-set manipulations. For several models (notably Qwen2.5 and OLMo3), CoT both increases Base accuracy and reduces the drop under \textsc{Nota}-as-answer, consistent with reduced reliance on specific answer options. However, this effect is not universal: Llama3.1 still shows a large gap between Base and \textsc{Nota}-as-answer even with CoT, indicating that correctness continues to depend on the presence of the original correct option. This residual gap is consistent with partially option-dependent CoT, in which plausible rationales are produced while some reliance on option matching remains.
SciQ exhibits a more consistent pattern across models.
Adding \textsc{Nota} as a distractor has only a small effect on accuracy under both Standard and CoT, suggesting that merely inserting an additional wrong option does not substantially disrupt model behaviour. In contrast, \textsc{Nota}-as-answer causes a pronounced drop across models, directly revealing reliance on the correct option's presence in the candidate set. Paired item-level tests confirm this asymmetry across model, dataset, and prompting conditions (Appendix~\ref{sec:statistical_validation}; Table~\ref{tab:nota_effect}).
CoT can raise Base accuracy and sometimes narrows this gap, but the remaining degradation indicates that option dependence is reduced rather than eliminated.
Overall, CoT mitigates parametric shortcuts \textit{to the extent that it reduces option dependence}. MathQA shows that CoT can improve robustness for some models, while SciQ confirms a strong and broadly shared failure mode: performance collapses when the ground-truth option is removed, and CoT only partially alleviates this. Therefore, \textsc{Nota} stress tests provide a necessary complement to standard leaderboard accuracy for diagnosing whether CoT gains reflect genuinely option-invariant reasoning.

\section{Conclusion}

In this paper, we introduced a task-centric framework that combines layer-wise, decoding-time, representation-based, and behavioural signals to characterise the demands that benchmarks place on LLMs.
Across 16 models and 24 benchmarks, we find that knowledge- and reasoning-centric tasks exhibit different internal processing patterns, and that these differences are not fully captured by aggregate task performance alone. At the behavioural level, knowledge performance scales more consistently than reasoning performance. Paired model-level tests detect a significant increase in reasoning scores under CoT. They do not detect a significant change in knowledge scores.
At the same time, the \textsc{Nota} perturbation analysis shows that improved performance under CoT does not necessarily imply option-invariant reasoning: models can remain strongly dependent on the presence of the original ground-truth answer.
Taken together, these results illustrate the value of analysing model behaviour along multiple complementary axes. Static layer-wise profiles characterise how task inputs are processed before generation. Decoding-time trajectories reveal differences in within-trajectory variability. Representation-based and perturbation analyses expose distinctions that may remain hidden in final task accuracy. Rather than treating knowledge and reasoning as disjoint capabilities, we provide a view of how their relative demands vary across tasks, models, and prompting conditions.

\section*{Code and Data Availability}

Our codebase for reproducing the figures and tables are available at \url{https://github.com/extraordinarylab/trace-kr}.
To ensure accessibility, all figures in this work utilise the colour-blind safe Okabe-Ito palette.

\section*{Limitations}
\label{sec:limitations}

Our study is limited to English benchmarks, and the probing, representation-similarity, and \textsc{Nota} analyses rely on multiple-choice answer structure. Cross-lingual differences in tokenisation and model coverage may change the observed footprints, while open-ended generation may exhibit different reasoning paths and shortcut behaviours. These settings define the scope of our conclusions and motivate future multilingual and open-ended evaluation.

\section*{Acknowledgments}

We are grateful to the action editor and the anonymous reviewers for their thoughtful feedback and valuable suggestions, which have helped improve the quality and clarity of this work.
This work was supported by UKRI Metascience-funded RISC project  (\mbox{APP111963}).
Eleni Vasilaki was supported by the UK Neuromorphic Computing Hardware Semiconductor IKC (Neuroware; UKRI2784), funded by EPSRC and Innovate UK.

\bibliography{tacl2021}
\bibliographystyle{acl_natbib}

\appendix

\clearpage

\section{Experiments Details}

\begin{table*}[t]
 \centering
 \resizebox{\linewidth}{!}{
 \begin{tabular}{llllr}
 \toprule
 \textbf{Dataset} & \textbf{Domain} & \textbf{Question Type} & \textbf{Metric} & \textbf{\# Examples} \\
 \midrule
 \rowcolor{AC!35} BioMixQA~\cite{soman2024biomedical} & Biomedical & Multiple-choice & Accuracy & 306 \\
 \rowcolor{AC!35} MedMCQA~\cite{pal2022medmcqa} & Biomedical & Multiple-choice & Accuracy & 4,183 \\
 \rowcolor{AC!35} MRI-MCQA~\cite{mri2025} & Biomedical & Multiple-choice & Accuracy & 563 \\
 \rowcolor{AC!35} PubMedQA~\cite{jin-etal-2019-pubmedqa} & Biomedical & Yes/No & Accuracy & 1,000 \\
 \rowcolor{AU!35} SciQ~\cite{welbl-etal-2017-crowdsourcing} & Science & Multiple-choice & Accuracy & 1,000 \\
 \rowcolor{AU!35} QASC~\cite{khot2020qasc} & Science & Multiple-choice & Accuracy & 926 \\
 \rowcolor{AU!35} AI2-ARC~\cite{clark2018think} & Science & Multiple-choice & Accuracy & 3,548 \\
 \rowcolor{AU!35} GPQA-Diamond~\cite{rein2024gpqa} & Science & Multiple-choice & Accuracy & 198 \\
 \rowcolor{TNG!35} MusicTrivia~\cite{music2025} & Music Knowledge & Multiple-choice & Accuracy & 512 \\
 \rowcolor{IE!35} RACE~\cite{lai-etal-2017-race} & Reading Comprehension & Multiple-choice & Accuracy & 4,934 \\
 \rowcolor{ST!35} TruthfulQA~\cite{lin-etal-2022-truthfulqa} & Factuality Detection & Multiple-choice & Accuracy & 684 \\
 \rowcolor{ST!35} HaluEval~\cite{li-etal-2023-halueval} & Factuality Detection & Yes/No & Accuracy & 30,000 \\
 \rowcolor{QU!35} CommonsenseQA~\cite{talmor-etal-2019-commonsenseqa} & Commonsense Reasoning & Multiple-choice & Accuracy & 1,221 \\
 \rowcolor{QU!35} Winogrande~\cite{sakaguchi2021winogrande} & Commonsense Reasoning & Multiple-choice & Accuracy & 1,267 \\
 \rowcolor{QU!35} PIQA~\cite{bisk2020piqa} & Commonsense Reasoning & Multiple-choice & Accuracy & 1,838 \\
 \rowcolor{QU!35} SIQA~\cite{sap-etal-2019-social} & Commonsense Reasoning & Multiple-choice & Accuracy & 1,954 \\
 \rowcolor{NE!35} MMLU-Pro~\cite{wang2024mmlu} & Multidisciplinary Knowledge & Multiple-choice & Accuracy & 12,032 \\
 \rowcolor{NS!35} MathQA~\cite{amini-etal-2019-mathqa} & Mathematics & Multiple-choice & Accuracy & 2,985 \\
 \rowcolor{NS!35} GSM8K~\cite{cobbe2021training} & Mathematics & Free-form & Exact Match & 1,319 \\
 \rowcolor{NS!35} AIME~\cite{aime25} & Mathematics & Free-form & Exact Match & 120 \\
 \rowcolor{NS!35} Math500~\cite{hendrycks2021measuring} & Mathematics & Free-form & Exact Match & 500 \\
 \rowcolor{NS!35} AMC~\cite{amc2025} & Mathematics & Free-form & Exact Match & 134 \\
 \rowcolor{YW!35} CoinFlip~\cite{wei2022chain} & Symbolic Reasoning & Yes/No & Accuracy & 3,333 \\
 \rowcolor{WP!35} LogiQA~\cite{liu2020logiqa} & Logical Reasoning & Multiple-choice & Accuracy & 651 \\
 \bottomrule
 \end{tabular}
 }
 \caption{Statistics of datasets used for knowledge and reasoning analysis, grouped by domain.}
 \label{tab:dataset_stats}
\end{table*}

\begin{table*}[t]
\centering
\small
\begin{tabular}{lllll}
\toprule
\textbf{Models} & \textbf{Weights} & \textbf{Model Link} & \textbf{License} & \textbf{Release Time} \\
\toprule
\rowcolor{AC!35} Llama3-8B & Dense & \href{https://huggingface.co/meta-llama/Meta-Llama-3-8B-Instruct}{meta-llama/Meta-Llama-3-8B-Instruct} & Llama 3 Community & Apr 2024 \\
\rowcolor{AC!35} Llama3.1-8B & Dense & \href{https://huggingface.co/meta-llama/Llama-3.1-8B-Instruct}{meta-llama/Llama-3.1-8B-Instruct} & Llama 3.1 Community & Jul 2024 \\
\rowcolor{AU!35} Qwen2.5-3B & Dense & \href{https://huggingface.co/Qwen/Qwen2.5-3B-Instruct}{Qwen/Qwen2.5-3B-Instruct} & Apache 2.0 & Sep 2024 \\
\rowcolor{AU!35} Qwen2.5-7B & Dense & \href{https://huggingface.co/Qwen/Qwen2.5-7B-Instruct}{Qwen/Qwen2.5-7B-Instruct} & Apache 2.0 & Sep 2024 \\
\rowcolor{AU!35} Qwen2.5-14B & Dense & \href{https://huggingface.co/Qwen/Qwen2.5-14B-Instruct}{Qwen/Qwen2.5-14B-Instruct} & Apache 2.0 & Sep 2024 \\
\rowcolor{TNG!35} Qwen3-8B & Dense & \href{https://huggingface.co/Qwen/Qwen3-8B}{Qwen/Qwen3-8B} & Apache 2.0 & Apr 2025 \\
\rowcolor{TNG!35} Qwen3-30B-A3B & MoE & \href{https://huggingface.co/Qwen/Qwen3-30B-A3B-Instruct-2507}{Qwen/Qwen3-30B-A3B-Instruct-2507} & Apache 2.0 & Jul 2025 \\
\rowcolor{IE!35} Gemma-7B & Dense & \href{https://huggingface.co/google/gemma-7b-it}{google/gemma-7b-it} & Gemma Terms of Use & Feb 2024 \\
\rowcolor{ST!35} Gemma-2-2B & Dense & \href{https://huggingface.co/google/gemma-2-2b-it}{google/gemma-2-2b-it} & Gemma Terms of Use & Jul 2024 \\
\rowcolor{ST!35} Gemma-2-9B & Dense & \href{https://huggingface.co/google/gemma-2-9b-it}{google/gemma-2-9b-it} & Gemma Terms of Use & Jun 2024 \\
\rowcolor{QU!35} Gemma-3-4B & Dense & \href{https://huggingface.co/google/gemma-3-4b-it}{google/gemma-3-4b-it} & Gemma Terms of Use & Mar 2025 \\
\rowcolor{QU!35} Gemma-3-12B & Dense & \href{https://huggingface.co/google/gemma-3-12b-it}{google/gemma-3-12b-it} & Gemma Terms of Use & Mar 2025 \\
\rowcolor{QU!35} Gemma-3-27B & Dense & \href{https://huggingface.co/google/gemma-3-27b-it}{google/gemma-3-27b-it} & Gemma Terms of Use & Mar 2025 \\
\rowcolor{NE!35} GLM-4-9B & Dense & \href{https://huggingface.co/zai-org/glm-4-9b-chat-hf}{zai-org/glm-4-9b-chat-hf} & Apache 2.0 & Oct 2024 \\
\rowcolor{NS!35} Olmo-3-7B & Dense & \href{https://huggingface.co/allenai/Olmo-3-7B-Instruct}{allenai/Olmo-3-7B-Instruct} & Apache 2.0 & Nov 2025 \\
\rowcolor{YW!35} GPT-OSS-20B & MoE & \href{https://huggingface.co/openai/gpt-oss-20b}{openai/gpt-oss-20b} & Apache 2.0 & Aug 2025 \\
\bottomrule
\end{tabular}
\caption{Overview of the models evaluated in this work, including their HuggingFace links, licensing terms, and release time. Models from the same family share the same background colour.}
\label{tab:models_details}
\end{table*}

\subsection{Datasets}
\label{sec:datasets}

\subsubsection{Benchmark Taxonomy}
\label{sec:benchmark_taxonomy}

Following the operational criteria in the main text, the knowledge-centric group comprises BioMixQA, MedMCQA, MRI-MCQA, SciQ, QASC, MusicTrivia, TruthfulQA, and HaluEval. These benchmarks predominantly test retrieval or verification of biomedical, scientific, cultural, or factual information. The reasoning-centric group comprises AI2-ARC, GPQA-Diamond, RACE, PubMedQA, CommonsenseQA, Winogrande, PIQA, SIQA, MMLU-Pro, MathQA, GSM8K, AIME, MATH500, AMC, CoinFlip, and LogiQA. Their items predominantly require composition, inference from a supplied context, calculation, or state tracking. This grouping records the dominant demand of a benchmark rather than asserting that every item is pure: GPQA and MMLU-Pro require substantial domain knowledge, while SciQ and biomedical QA can require local inference. These mixed cases are precisely why we analyse a graded spectrum and report deviations from the group-level trend.

\paragraph{BioMixQA} \cite{soman2024biomedical} is a biomedical MCQA dataset created to evaluate Knowledge Graph-based Retrieval-Augmented Generation (KG-RAG) frameworks.

\paragraph{MedMCQA} \cite{pal2022medmcqa} is a large-scale MCQA dataset with over 194,000 questions from Indian medical entrance exams, testing complex reasoning across 21 medical subjects.

\paragraph{MRI-MCQA} \cite{mri2025} is a specialised MCQA benchmark covering Magnetic Resonance Imaging (MRI) topics such as cardiac imaging, pulse sequences, and physics.

\paragraph{PubMedQA} \cite{jin-etal-2019-pubmedqa} is a biomedical QA dataset requiring yes/no/maybe answers to research questions based on PubMed abstracts, testing the understanding of quantitative results.

\paragraph{SciQ} \cite{welbl-etal-2017-crowdsourcing} is a crowdsourced dataset of 13,679 multiple-choice science questions in Physics, Chemistry, and Biology, most of which include a supporting text with evidence.

\paragraph{QASC} \cite{khot2020qasc} is a grade-school science dataset with nearly 10,000 questions designed to test multi-hop reasoning by composing facts.

\paragraph{AI2-ARC} \cite{clark2018think} is a dataset of 7,787 grade-school science questions. Its Challenge Set of 2,590 questions is designed to require complex reasoning beyond simple retrieval or word association.

\paragraph{GPQA-Diamond} \cite{rein2024gpqa} is a Google-Proof dataset of 448 challenging science questions created by PhD-level experts, designed to be difficult even with internet access.

\paragraph{MusicTrivia} \cite{music2025} is an MCQA dataset testing a wide range of musical knowledge, from classical composers and periods to modern pop/rock artists and general music history.

\paragraph{RACE} \cite{lai-etal-2017-race} is a large-scale reading comprehension dataset with nearly 100,000 questions sourced from English exams for middle and high school students.

\paragraph{TruthfulQA} \cite{lin-etal-2022-truthfulqa} is a benchmark designed to measure a model's truthfulness using 817 adversarial questions that probe common human misconceptions.

\paragraph{HaluEval} \cite{li-etal-2023-halueval} is a large-scale benchmark for evaluating an LLM's ability to recognize and avoid generating factually incorrect or unverifiable content (hallucinations).

\paragraph{CommonsenseQA} \cite{talmor-etal-2019-commonsenseqa} is a challenging MCQA dataset designed to evaluate commonsense reasoning ability beyond simple fact retrieval.

\paragraph{Winogrande} \cite{sakaguchi2021winogrande} is a large-scale and challenging dataset for commonsense reasoning, designed as a more robust successor to the Winograd Schema Challenge.

\paragraph{PIQA} \cite{bisk2020piqa} is a benchmark for physical commonsense, requiring models to choose the more sensible of two solutions to achieve a practical, real-world goal.

\paragraph{SIQA} \cite{sap-etal-2019-social} is an MCQA benchmark for testing social and emotional commonsense by requiring reasoning about people's motivations and actions in social situations.

\paragraph{MMLU-Pro} \cite{wang2024mmlupro} is an enhanced version of MMLU that increases difficulty with more complex questions and expands from four to ten multiple-choice options for a more robust evaluation.

\paragraph{MathQA} \cite{amini-etal-2019-mathqa} is a large-scale dataset for math word problems where each problem is paired with an operational program, promoting interpretable, step-by-step solutions.

\paragraph{GSM8K} \cite{cobbe2021training} is a dataset of 8,500 high-quality grade school math word problems designed to evaluate multi-step reasoning, with solutions provided as natural language explanations.

\paragraph{AIME} \cite{aime25} is a benchmark of problems from the American Invitational Mathematics Examination, a challenging high school competition used to evaluate advanced mathematical reasoning.

\paragraph{MATH500} \cite{hendrycks2021measuring} is a benchmark of thousands of competition math problems from pre-algebra to calculus, used to evaluate mathematical problem-solving via step-by-step solutions.

\paragraph{AMC} \cite{amc2025} consists of problems from the American Mathematics Competitions, used to test pre-collegiate proficiency in algebra, geometry, number theory, and combinatorics.

\paragraph{CoinFlip} \cite{wei2022chain} is a symbolic reasoning benchmark that tests a model's ability to track the binary state of an object through a sequence of actions.

\paragraph{LogiQA} \cite{liu2020logiqa} is a reading comprehension benchmark from civil service exams that tests formal logical reasoning (e.g., categorical, conditional) based on provided premises.

\subsection{Models}
\label{sec:models}

\begin{figure*}[t]
    \centering
    \includegraphics[width=1\linewidth]{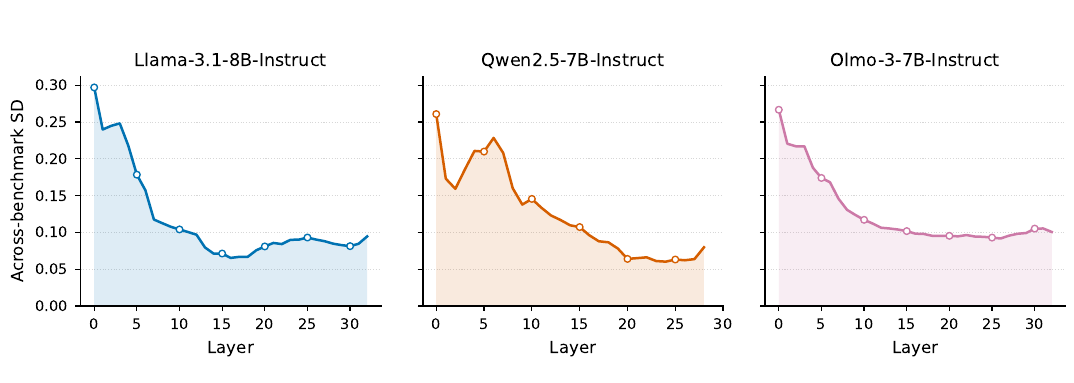}
    \caption{
    Cross-benchmark variability in adjusted cosine similarity for the Llama, Qwen, and OLMo model families. The plotted value is the population standard deviation of those four benchmark means. Dispersion is greatest in early layers and decreases toward a lower, relatively stable level in middle and later layers.
    }
    \label{fig:separability}
\end{figure*}

\begin{figure*}[t]
    \begin{tcolorbox}[
        enhanced,
        colback=white,
        colframe=black,
        arc=3mm,
        boxrule=0.5pt,
        title=Prompt Template (Retrieve $\rightarrow$ Reason).
    ]
    Identify key facts and technical principles related to the following question. Based only on the knowledge retrieved above, analyze the following question. The last line of your response should be of the following format: 'ANSWER: [LETTER]' (without quotes) where [LETTER] is one of {letters}.

    \vspace{1ex}
    \textbf{Question}: \{question\}

    \textbf{Choices}: \{choices\}

    \end{tcolorbox}
    \caption{Prompt used for decoding-time trajectory analysis.}
    \label{fig:trajectory_prompt}
\end{figure*}

Table~\ref{tab:models_details} provides an overview of the open-source LLMs evaluated in this work. Models from the same family share a common background colour.

\begin{figure*}[t]
\centering
\begin{minipage}[t]{0.48\textwidth}
\begin{tcolorbox}[
    enhanced,
    colback=white,
    colframe=black,
    arc=2mm,
    boxrule=0.5pt,
    title={(a) Prompt for knowledge-centric datasets}
]
\vspace{0.5ex}
\textbf{Question}: \{question\}

\textbf{Choices}: \{choices\}

\vspace{1ex}
Respond with one letter. The answer is \_
\end{tcolorbox}
\end{minipage}\hfill
\begin{minipage}[t]{0.48\textwidth}
\begin{tcolorbox}[
    enhanced,
    colback=white,
    colframe=black,
    arc=2mm,
    boxrule=0.5pt,
    title={(b) Prompt for reasoning-centric datasets}
]
\vspace{0.5ex}
Please enclose your thinking process in \texttt{<think></think>} and the final answer in \texttt{<final\_answer></final\_answer>}.

\vspace{1ex}
\textbf{Question}: \{question\}

\textbf{Choices}: \{choices\}

\vspace{1ex}
Respond with one letter. The answer is \texttt{\_ <think></think><final\_answer>}
\end{tcolorbox}
\end{minipage}

\caption{Prompt templates for extracting representations under the multiple-choice setting. We query the LLM with the appropriate template and extract the last-token representation from each layer as input to the linear probing classifiers.}
\label{fig:probing_prompt_mcp}
\end{figure*}

\subsection{Prompts}

Our prompt templates are designed to (i) standardise the interface across benchmarks for representation extraction, (ii) avoid confounds from variable-length decoding. Importantly, we do \textit{not} generate free-form outputs for probing. Instead, we run a single forward pass on the full prompt and extract the hidden state at the last input token from each layer, which serves as the feature for the layer-wise linear classifiers.

\begin{table}[t]
\centering
\setlength{\tabcolsep}{4pt}

\resizebox{\linewidth}{!}{
\begin{tabular}{c cc cc cc}
\toprule
\multirow{2}{*}{\textbf{Metric}}
& \multicolumn{2}{c}{\textbf{Lower}}
& \multicolumn{2}{c}{\textbf{Middle}}
& \multicolumn{2}{c}{\textbf{Upper}} \\
\cmidrule(lr){2-3}
\cmidrule(lr){4-5}
\cmidrule(lr){6-7}
& $q$ & $\delta$
& $q$ & $\delta$
& $q$ & $\delta$ \\
\midrule
Attention Entropy
& \textbf{$<$.001} & .42
& .055 & .14
& \textbf{$<$.001} & .45 \\
Activation Sparsity
& .999 & $-$.27
& .999 & $-$.07
& .205 & .13 \\
\bottomrule
\end{tabular}
}

\caption{
Correct-incorrect differences in within-trajectory variability across representative lower, middle, and upper layers. $q$ denotes the Benjamini-Hochberg adjusted $p$-value \cite{benjamini_hochberg_1995}, with correction across the three layer-wise comparisons within each metric. $\delta$ denotes Cliff's delta \cite{cliff1993dominance}; positive values indicate greater within-trajectory variability for incorrect than correct generations. \textbf{Bold} values indicate $q<.05$.
}
\label{tab:trajectory_variability}
\end{table}

\begin{table}[t]
\centering
\setlength{\tabcolsep}{4pt}

\resizebox{\linewidth}{!}{
\begin{tabular}{ccccc}
\toprule
\begin{tabular}[c]{@{}c@{}}\textbf{Capability}\end{tabular}
& \begin{tabular}[c]{@{}c@{}}\textbf{Median}\\\textbf{change}\end{tabular}
& \begin{tabular}[c]{@{}c@{}}\textbf{Models}\\\textbf{improved}\end{tabular}
& \begin{tabular}[c]{@{}c@{}}\textbf{p-value}\end{tabular}
& \begin{tabular}[c]{@{}c@{}}\textbf{Effect}\\\textbf{size}\end{tabular} \\
\midrule
Reasoning
& +.036
& 14/16
& .008
& +.74 \\

Knowledge
& +.001
& 9/16
& .597
& +.16 \\
\bottomrule
\end{tabular}
}

\caption{
Effect of CoT prompting on knowledge and reasoning. Median change summarises paired score differences from Standard to CoT; counts show improved checkpoints out of 16. We report two-sided Wilcoxon p-values \cite{wilcoxon1945individual} and matched-pairs rank-biserial effect sizes \cite{kerby2014simple}.
}
\label{tab:cot_effect}
\end{table}
\begin{table}[t]
\centering
\small
\setlength{\tabcolsep}{4pt}
\begin{tabular}{@{}ccc@{}}
\toprule
\begin{tabular}[c]{@{}c@{}}\textbf{Perturbation}\end{tabular}
& \begin{tabular}[c]{@{}c@{}}\textbf{Median}\\\textbf{accuracy drop}\end{tabular}
& \begin{tabular}[c]{@{}c@{}}\textbf{Significant}\\\textbf{comparisons}\end{tabular} \\
\midrule
\textsc{Nota}-as-answer
& .346
& 24/24 \\

\textsc{Nota}-as-distractor
& .012
& 3/24 \\
\bottomrule
\end{tabular}

\caption{
Robustness to \textsc{Nota} perturbations across model, dataset, and prompting conditions. Accuracy drops are medians relative to Base. Counts show significant comparisons ($q<.05$) out of 24 per perturbation, after Benjamini-Hochberg correction across all 48 exact McNemar tests \cite{mcnemar1947note}.
}
\label{tab:nota_effect}
\end{table}

\subsection{Statistical Validation}
\label{sec:statistical_validation}

We complement the descriptive comparisons in RQ2 with paired non-parametric tests targeting the corresponding experimental units. For the CoT analysis, comparisons are paired across model checkpoints. For the \textsc{Nota} analysis, comparisons are paired at the item level within each model, dataset, and prompting condition.

\paragraph{CoT prompting.}~To assess whether the shifts in Figure~\ref{fig:landscape} are systematic across models, we compare the Standard and CoT scores separately for the knowledge and reasoning dimensions using two-sided Wilcoxon signed-rank tests \citep{wilcoxon1945individual}. The model checkpoint is treated as the paired unit; the \textit{Think} and \textit{No-Think} configurations of Qwen3-8B are aggregated within the shared checkpoint before testing. We report the median paired change, the number of models with a positive change, and the matched-pairs rank-biserial correlation $r_{\mathrm{rb}}$ \citep{kerby2014simple} as an effect size. Table~\ref{tab:cot_effect} shows a systematic positive shift in reasoning, whereas the corresponding shift in knowledge is not supported.

\paragraph{\textsc{Nota} perturbations.}~For Figures~\ref{fig:mathqa_dumbbell} and~\ref{fig:sciq_dumbbell}, we test changes in item-level correctness between each Base condition and its corresponding \textsc{Nota} perturbation using two-sided exact McNemar tests \citep{mcnemar1947note}. Tests are conducted separately for each model $\times$ dataset $\times$ prompting condition, yielding 24 comparisons for \textsc{Nota}-as-answer and 24 for \textsc{Nota}-as-distractor. We control the false discovery rate across the resulting 48 tests using the Benjamini-Hochberg procedure \citep{benjamini_hochberg_1995}. As summarised in Table~\ref{tab:nota_effect}, the \textsc{Nota}-as-answer effect is consistently detected across conditions, whereas the \textsc{Nota}-as-distractor effect is comparatively limited.

\begin{sidewaystable*}[t]
\centering
\resizebox{\textwidth}{!}{
\begin{tabular}{lcccccccccccccccccc}
\toprule
\multirow{2}{*}{Model} & \multicolumn{2}{c}{MRI-MCQA (0-shot)} & \multicolumn{2}{c}{BioMixQA (0-shot)} & \multicolumn{2}{c}{SciQ (0-shot)} & \multicolumn{2}{c}{QASC (0-shot)} & \multicolumn{2}{c}{MusicTrivia (0-shot)} & \multicolumn{2}{c}{TruthfulQA (0-shot)} & \multicolumn{2}{c}{MedMCQA (0-shot)} \\
\cmidrule(lr){2-3} \cmidrule(lr){4-5} \cmidrule(lr){6-7} \cmidrule(lr){8-9} \cmidrule(lr){10-11} \cmidrule(lr){12-13} \cmidrule(lr){14-15}
& Std. & CoT & Std. & CoT & Std. & CoT & Std. & CoT & Std. & CoT & Std. & CoT & Std. & CoT \\
\midrule
Llama3-8B-Instruct & 52.58 & 50.80 & 50.98 & 45.75 & 90.60 & 90.70 & 76.24 & 74.51 & 76.37 & 73.44 & 52.05 & 54.39 & 57.42 & 53.45 \\
Llama3.1-8B-Instruct & 54.88 & 55.77 & 49.67 & 49.67 & 91.60 & 93.30 & 79.59 & 80.78 & 73.83 & 76.76 & 55.12 & 59.94 & 56.87 & 59.12 \\
Qwen2.5-3B-Instruct & 49.38 & 48.31 & 36.60 & 37.91 & 91.70 & 92.00 & 72.03 & 74.30 & 71.48 & 71.29 & 59.94 & 57.02 & 48.24 & 47.12 \\
Qwen2.5-7B-Instruct & 55.95 & 56.48 & 39.87 & 40.52 & 94.70 & 94.40 & 79.59 & 82.07 & 80.47 & 79.49 & 69.88 & 64.04 & 55.22 & 55.13 \\
Qwen2.5-14B-Instruct & 60.04 & 63.23 & 58.82 & 57.19 & 95.80 & 95.90 & 82.94 & 85.85 & 85.55 & 86.52 & 78.07 & 73.39 & 60.41 & 60.27 \\
Qwen3-8B (No-Think) & 67.14 & 54.71 & 57.52 & 44.77 & 96.90 & 95.20 & 86.93 & 83.26 & 84.57 & 82.42 & 74.85 & 73.25 & 65.34 & 57.14 \\
Qwen3-8B (Think) & 67.50 & 70.69 & 57.52 & 57.19 & 96.50 & 95.40 & 86.50 & 82.51 & 84.96 & 81.64 & 73.68 & 72.66 & 65.31 & 56.56 \\
Gemma-7B-Instruct & 40.14 & 25.93 & 38.89 & 14.71 & 85.50 & 64.30 & 67.71 & 46.11 & 57.03 & 37.11 & 31.87 & 23.50 & 39.88 & 28.74 \\
Gemma-2-2B-Instruct & 41.74 & 36.23 & 35.29 & 37.25 & 88.50 & 88.30 & 67.39 & 67.39 & 67.19 & 62.11 & 46.93 & 50.73 & 41.55 & 39.97 \\
Gemma-2-9B-Instruct & 54.35 & 53.64 & 49.02 & 46.41 & 94.90 & 93.90 & 80.89 & 80.35 & 82.23 & 79.88 & 72.81 & 73.83 & 54.46 & 53.69 \\
Gemma-3-4B-Instruct & 41.39 & 43.34 & 34.64 & 32.35 & 86.80 & 88.50 & 67.49 & 72.35 & 66.80 & 70.90 & 51.02 & 48.98 & 43.39 & 46.21 \\
Gemma-3-12B-Instruct & 56.13 & 61.63 & 52.29 & 49.35 & 93.50 & 94.50 & 80.56 & 82.29 & 84.18 & 83.59 & 68.57 & 65.94 & 55.51 & 58.81 \\
Gemma-3-27B-Instruct & 60.92 & 63.94 & 50.98 & 52.61 & 95.30 & 94.90 & 82.51 & 83.80 & 85.35 & 86.33 & 75.15 & 72.37 & 61.20 & 64.12 \\
GLM-4-9B-Instruct & 45.47 & 44.58 & 43.79 & 37.25 & 91.10 & 90.80 & 74.84 & 71.60 & 73.63 & 74.41 & 56.43 & 53.80 & 46.98 & 45.21 \\
Olmo-3-7B-Instruct & 47.42 & 53.11 & 34.64 & 48.69 & 88.80 & 92.30 & 65.77 & 75.16 & 67.97 & 70.12 & 60.09 & 56.58 & 41.07 & 47.86 \\
GPT-OSS-20B & 63.06 & 67.32 & \textbf{61.76} & 62.09 & 51.50 & 81.00 & 49.68 & 72.79 & 60.94 & 74.80 & 42.11 & 64.77 & 50.25 & 57.04 \\
Qwen3-30B-A3B & \textbf{71.94} & \textbf{74.07} & 59.48 & \textbf{64.38} & \textbf{97.30} & \textbf{97.30} & \textbf{89.09} & \textbf{88.44} & \textbf{87.50} & \textbf{87.70} & \textbf{78.07} & \textbf{75.58} & \textbf{70.95} & \textbf{71.00} \\
\midrule
\multirow{2}{*}{Model} & \multicolumn{2}{c}{HaluEval-D (6-shot)} & \multicolumn{2}{c}{HaluEval-Q (7-shot)} & \multicolumn{2}{c}{HaluEval-S (3-shot)} \\
\cmidrule(lr){2-3} \cmidrule(lr){4-5} \cmidrule(lr){6-7}
& Std. & CoT & Std. & CoT & Std. & CoT \\
\midrule
Llama3-8B-Instruct & 82.10 & 93.94 & 78.98 & 82.69 & 71.74 & 83.86 \\
Llama3.1-8B-Instruct & 80.22 & 81.74 & 68.33 & 73.28 & 65.52 & 76.48 \\
Qwen2.5-3B-Instruct & 52.19 & 50.61 & 49.14 & 49.65 & 45.45 & 45.66 \\
Qwen2.5-7B-Instruct & 90.87 & 90.27 & 67.33 & 67.52 & 58.75 & 53.73 \\
Qwen2.5-14B-Instruct & 94.04 & 93.89 & 88.39 & 88.11 & 84.90 & 85.40 \\
Qwen3-8B (No-Think) & 79.56 & 89.30 & 84.56 & 87.28 & 69.14 & 73.49 \\
Qwen3-8B (Think) & 79.89 & 88.86 & 84.90 & 87.92 & 69.21 & 73.47 \\
Gemma-7B-Instruct & 58.37 & 59.42 & 64.03 & 65.36 & \textbf{96.04} & \textbf{96.95} \\
Gemma-2-2B-Instruct & 56.35 & 55.37 & 28.41 & 28.82 & 44.19 & 45.54 \\
Gemma-2-9B-Instruct & 71.95 & 71.89 & 70.73 & 73.21 & 68.72 & 71.81 \\
Gemma-3-4B-Instruct & 67.46 & 67.31 & 46.88 & 48.27 & 57.00 & 59.34 \\
Gemma-3-12B-Instruct & 79.11 & 79.29 & 47.00 & 43.86 & 58.08 & 57.02 \\
Gemma-3-27B-Instruct & 81.13 & 81.21 & 44.10 & 41.07 & 71.21 & 70.65 \\
GLM-4-9B-Instruct & 73.92 & 74.26 & 34.83 & 35.71 & 53.14 & 55.80 \\
Olmo-3-7B-Instruct & 67.77 & 67.27 & 50.13 & 49.88 & 52.48 & 52.93 \\
GPT-OSS-20B & \textbf{98.20} & \textbf{98.62} & \textbf{95.38} & \textbf{95.95} & 97.30 & 97.42 \\
Qwen3-30B-A3B & 77.69 & 77.97 & 83.13 & 83.08 & 69.52 & 68.83 \\
\bottomrule
\end{tabular}
}
\caption{
    Performance of all models across knowledge-based benchmarks. For each dataset, we report accuracy under both standard (Std.) and CoT prompting. The best-performing model in each setting is highlighted in \textbf{bold}.
}
\label{tab:full_knowledge_results_appendix}
\end{sidewaystable*}

\begin{sidewaystable*}[t]
\centering
\resizebox{\textwidth}{!}{
\begin{tabular}{lcccccccccccccccc}
\toprule
\multirow{2}{*}{Model} & \multicolumn{2}{c}{GSM8K (8-shot)} & \multicolumn{2}{c}{MathQA (0-shot)} & \multicolumn{2}{c}{MATH500 (0-shot)} & \multicolumn{2}{c}{AMC (0-shot)} & \multicolumn{2}{c}{AIME22 (0-shot)} & \multicolumn{2}{c}{AIME23 (0-shot)} & \multicolumn{2}{c}{AIME24 (0-shot)} & \multicolumn{2}{c}{AIME25 (0-shot)} \\
\cmidrule(lr){2-3} \cmidrule(lr){4-5} \cmidrule(lr){6-7} \cmidrule(lr){8-9} \cmidrule(lr){10-11} \cmidrule(lr){12-13} \cmidrule(lr){14-15} \cmidrule(lr){16-17}
& Std. & CoT & Std. & CoT & Std. & CoT & Std. & CoT & Std. & CoT & Std. & CoT & Std. & CoT & Std. & CoT \\
\midrule
Llama3-8B-Instruct & 74.75 & 75.06 & 54.84 & 55.28 & 23.20 & 23.80 & 5.22 & 7.46 & 0.00 & 0.00 & 0.00 & 0.00 & 0.00 & 3.33 & 0.00 & 0.00 \\
Llama3.1-8B-Instruct & 78.77 & 82.26 & 54.17 & 69.45 & 47.40 & 50.20 & 12.69 & 16.42 & 6.67 & 0.00 & 3.33 & 3.33 & 3.33 & 6.67 & 0.00 & 0.00 \\
Qwen2.5-3B-Instruct & 79.08 & 81.65 & 77.79 & 77.25 & 65.80 & 65.60 & 27.61 & 38.06 & 3.33 & 3.33 & 6.67 & 3.33 & 16.67 & 3.33 & 0.00 & 3.33 \\
Qwen2.5-7B-Instruct & 86.66 & 88.48 & 38.63 & 83.28 & 76.40 & 73.40 & 42.54 & 43.28 & 16.67 & 6.67 & 13.33 & 10.00 & 16.67 & 10.00 & 20.00 & 13.33 \\
Qwen2.5-14B-Instruct & 90.75 & 90.14 & 48.54 & 85.16 & 78.60 & 80.40 & 48.51 & 51.49 & 6.67 & 10.00 & 10.00 & 13.33 & 10.00 & 20.00 & 16.67 & 13.33 \\
Qwen3-8B (No-Think) & 88.40 & 90.52 & 53.74 & 89.65 & 22.00 & 24.40 & 8.96 & 80.60 & 3.33 & 0.00 & 0.00 & 0.00 & 0.00 & 0.00 & 3.33 & 3.33 \\
Qwen3-8B (Think) & 89.31 & 91.13 & 68.94 & 89.78 & 22.40 & 22.00 & 8.21 & \textbf{89.55} & 0.00 & 10.00 & 3.33 & 0.00 & 0.00 & 0.00 & 0.00 & 3.33 \\
Gemma-7B-Instruct & 16.68 & 24.72 & 20.13 & 23.25 & 6.20 & 5.00 & 2.24 & 0.00 & 0.00 & 0.00 & 0.00 & 0.00 & 0.00 & 0.00 & 0.00 & 0.00 \\
Gemma-2-2B-Instruct & 39.50 & 37.76 & 20.70 & 33.13 & 21.00 & 21.00 & 0.00 & 2.24 & 0.00 & 0.00 & 0.00 & 0.00 & 0.00 & 0.00 & 0.00 & 0.00 \\
Gemma-2-9B-Instruct & 79.45 & 81.80 & 28.07 & 61.74 & 49.20 & 50.80 & 13.43 & 10.45 & 0.00 & 0.00 & 3.33 & 6.67 & 0.00 & 6.67 & 0.00 & 0.00 \\
Gemma-3-4B-Instruct & 81.35 & 82.49 & 25.80 & 73.63 & 73.20 & 76.60 & 37.31 & 38.81 & 3.33 & 10.00 & 6.67 & 10.00 & 6.67 & 20.00 & 10.00 & 16.67 \\
Gemma-3-12B-Instruct & 89.23 & 89.16 & 32.19 & 78.76 & 84.20 & 83.60 & 52.98 & 52.24 & 13.33 & 26.67 & 13.33 & 13.33 & 23.33 & 23.33 & 20.00 & 20.00 \\
Gemma-3-27B-Instruct & 92.65 & 91.66 & 46.73 & 85.29 & 87.80 & 88.40 & 60.45 & 58.96 & 16.67 & 16.67 & 20.00 & 13.33 & 26.67 & 30.00 & 26.67 & 23.33 \\
GLM-4-9B-Instruct & 65.73 & 64.75 & 28.71 & 50.22 & 28.40 & 27.40 & 7.46 & 5.97 & 0.00 & 0.00 & 0.00 & 3.33 & 0.00 & 0.00 & 0.00 & 0.00 \\
Olmo-3-7B-Instruct & 88.40 & 88.32 & 33.03 & 81.68 & 79.00 & 78.00 & 44.78 & 50.00 & 16.67 & 23.33 & 13.33 & 13.33 & 16.67 & 13.33 & 20.00 & 23.33 \\
GPT-OSS-20B & \textbf{94.54} & \textbf{93.78} & 75.08 & 78.16 & 79.60 & 76.20 & 18.66 & 20.15 & 13.33 & 6.67 & 10.00 & 6.67 & 16.67 & 20.00 & 16.67 & 20.00 \\
Qwen3-30B-A3B & 93.63 & 93.56 & \textbf{90.28} & \textbf{90.32} & \textbf{95.60} & \textbf{95.60} & \textbf{71.64} & 77.61 & \textbf{73.33} & \textbf{83.33} & \textbf{80.00} & \textbf{76.67} & \textbf{83.33} & \textbf{73.33} & \textbf{73.33} & \textbf{73.33} \\
\bottomrule
\end{tabular}
}
\caption{
    Performance on Mathematical Reasoning benchmarks (Part 1). Scores are reported as accuracy. The number of few-shot examples used is in parentheses. The best-performing model in each setting is highlighted in \textbf{bold}.
}
\label{tab:reasoning_part1_math}
\end{sidewaystable*}

\begin{sidewaystable*}[t]
\centering
\begin{tabular}{lcccccccccc}
\toprule
\multirow{2}{*}{Model} & \multicolumn{2}{c}{CSQA (0-shot)} & \multicolumn{2}{c}{SiQA (0-shot)} & \multicolumn{2}{c}{PiQA (0-shot)} & \multicolumn{2}{c}{Wino. (0-shot)} & \multicolumn{2}{c}{LogiQA (0-shot)} \\
\cmidrule(lr){2-3} \cmidrule(lr){4-5} \cmidrule(lr){6-7} \cmidrule(lr){8-9} \cmidrule(lr){10-11}
& Std. & CoT & Std. & CoT & Std. & CoT & Std. & CoT & Std. & CoT \\
\midrule
Llama3-8B-Instruct & 74.53 & 71.58 & 71.85 & 68.94 & 80.63 & 77.75 & 59.75 & 62.12 & 43.63 & 41.47 \\
Llama3.1-8B-Instruct & 73.71 & 74.61 & 67.40 & 73.13 & 81.01 & 83.35 & 60.85 & 64.33 & 41.94 & 42.09 \\
Qwen2.5-3B-Instruct & 75.59 & 73.38 & 72.98 & 70.01 & 77.15 & 79.33 & 63.69 & 64.33 & 43.47 & 46.24 \\
Qwen2.5-7B-Instruct & 82.06 & 79.61 & 78.05 & 77.18 & 87.05 & 87.21 & 66.93 & 72.38 & 53.76 & 54.22 \\
Qwen2.5-14B-Instruct & 82.06 & 81.41 & 78.56 & 78.35 & 89.17 & 92.00 & 77.19 & 81.14 & 61.60 & 58.22 \\
Qwen3-8B (No-Think) & 83.13 & 81.98 & 77.69 & 76.71 & 90.32 & 88.47 & 80.27 & 78.85 & 51.31 & 42.70 \\
Qwen3-8B (Think) & 83.46 & 81.65 & 76.71 & 76.97 & 89.93 & 88.85 & 81.69 & 77.51 & \textbf{75.58} & 53.00 \\
Gemma-7B-Instruct & 68.96 & 40.87 & 64.38 & 42.99 & 71.76 & 51.36 & 56.20 & 37.81 & 34.41 & 16.74 \\
Gemma-2-2B-Instruct & 69.78 & 68.30 & 69.29 & 66.63 & 74.59 & 71.76 & 53.91 & 52.57 & 43.01 & 38.10 \\
Gemma-2-9B-Instruct & 80.34 & 78.46 & 75.03 & 73.23 & 87.05 & 84.22 & 72.45 & 71.11 & 51.92 & 49.62 \\
Gemma-3-4B-Instruct & 67.81 & 70.02 & 66.68 & 67.55 & 77.97 & 79.76 & 56.83 & 58.96 & 46.54 & 43.78 \\
Gemma-3-12B-Instruct & 78.21 & 79.28 & 75.33 & 76.82 & 86.72 & 87.70 & 69.69 & 71.19 & 58.06 & 58.53 \\
Gemma-3-27B-Instruct & 79.77 & 79.12 & 77.12 & 77.18 & 89.88 & 89.66 & 73.48 & 75.69 & 61.60 & 62.52 \\
GLM-4-9B-Instruct & 78.87 & 76.66 & 74.41 & 74.92 & 80.36 & 81.99 & 69.93 & 67.72 & 48.39 & 45.78 \\
Olmo-3-7B-Instruct & 72.24 & 75.35 & 67.30 & 69.96 & 75.35 & 80.25 & 57.62 & 63.93 & 39.63 & 43.78 \\
GPT-OSS-20B & 48.16 & 73.87 & 54.61 & 73.64 & 40.75 & 77.64 & 56.43 & 73.56 & 54.22 & 54.53 \\
Qwen3-30B-A3B & \textbf{83.95} & \textbf{83.70} & \textbf{78.97} & \textbf{79.48} & \textbf{93.47} & \textbf{93.14} & \textbf{84.69} & \textbf{85.48} & 76.34 & \textbf{76.04} \\
\bottomrule
\end{tabular}
\caption{
    Performance on General and Commonsense Reasoning benchmarks (Part 2). Scores are reported as accuracy. The number of few-shot examples used is in parentheses. The best-performing model is highlighted in \textbf{bold}.
}
\label{tab:reasoning_part2_commonsense}
\end{sidewaystable*}

\begin{sidewaystable*}[t]
\centering
\resizebox{\textwidth}{!}{
\begin{tabular}{lcccccccccccccc}
\toprule
\multirow{2}{*}{Model} & \multicolumn{2}{c}{ARC-Easy (0-shot)} & \multicolumn{2}{c}{ARC-Chal. (0-shot)} & \multicolumn{2}{c}{RACE (3-shot)} & \multicolumn{2}{c}{PubMedQA (0-shot)} & \multicolumn{2}{c}{MMLU-Pro (5-shot)} & \multicolumn{2}{c}{GPQA-Dia. (0-shot)} & \multicolumn{2}{c}{CoinFlip (8-shot)} \\
\cmidrule(lr){2-3} \cmidrule(lr){4-5} \cmidrule(lr){6-7} \cmidrule(lr){8-9} \cmidrule(lr){10-11} \cmidrule(lr){12-13} \cmidrule(lr){14-15}
& Std. & CoT & Std. & CoT & Std. & CoT & Std. & CoT & Std. & CoT & Std. & CoT & Std. & CoT \\
\midrule
Llama3-8B-Instruct & 87.84 & 87.88 & 77.99 & 81.14 & 49.56 & 65.08 & 63.00 & 81.80 & 22.75 & 39.63 & 29.29 & 32.32 & 67.45 & 72.07 \\
Llama3.1-8B-Instruct & 88.01 & 89.98 & 79.27 & 85.07 & 72.31 & 82.37 & 70.90 & \textbf{82.20} & 44.96 & 46.99 & 28.79 & 26.26 & 59.89 & 75.31 \\
Qwen2.5-3B-Instruct & 89.56 & 89.31 & 82.08 & 81.83 & 81.19 & 81.25 & 58.40 & 66.10 & 44.42 & 44.85 & 29.80 & 33.84 & 58.39 & 65.89 \\
Qwen2.5-7B-Instruct & 92.13 & 92.26 & 87.63 & 89.93 & 87.63 & 86.16 & 53.60 & 58.10 & 56.58 & 56.70 & 32.83 & 34.85 & 73.24 & 70.81 \\
Qwen2.5-14B-Instruct & 93.73 & 93.86 & 91.13 & 91.81 & 91.06 & 90.70 & 44.80 & 74.80 & 64.66 & 64.83 & 32.83 & 40.91 & 71.83 & 84.01 \\
Qwen3-8B (No-Think) & 94.32 & 93.22 & 93.69 & 90.61 & 88.53 & 90.44 & 65.70 & 79.40 & 63.03 & 73.90 & 18.69 & 56.06 & 78.40 & 85.45 \\
Qwen3-8B (Think) & 94.49 & 93.06 & 93.77 & 91.04 & 89.34 & 90.42 & 65.60 & 68.70 & 63.07 & 78.05 & 17.68 & 58.08 & \textbf{86.35} & 85.45 \\
Gemma-7B-Instruct & 81.19 & 57.62 & 69.54 & 47.01 & 44.10 & 28.28 & 72.10 & 73.60 & 9.58 & 9.99 & 28.28 & 17.68 & 14.01 & 2.46 \\
Gemma-2-2B-Instruct & 84.60 & 85.90 & 72.01 & 75.34 & 74.79 & 61.49 & 39.50 & 76.50 & 24.28 & 25.32 & 30.30 & 17.17 & 42.24 & 35.01 \\
Gemma-2-9B-Instruct & 92.00 & 92.63 & 88.05 & 89.08 & 87.41 & 86.58 & 41.60 & 79.00 & 50.46 & 50.65 & 30.30 & 26.26 & 36.93 & 42.93 \\
Gemma-3-4B-Instruct & 85.06 & 86.41 & 74.74 & 80.72 & 76.57 & 76.83 & 55.20 & 78.30 & 42.19 & 41.71 & 33.84 & 24.75 & 50.02 & 52.69 \\
Gemma-3-12B-Instruct & 92.42 & 93.35 & 88.23 & 91.30 & 87.98 & 87.80 & 49.40 & 77.40 & 60.44 & 59.98 & 32.83 & 37.37 & 78.46 & 79.90 \\
Gemma-3-27B-Instruct & 93.81 & 93.86 & 91.04 & 92.15 & 88.95 & 89.48 & 58.20 & 80.50 & 67.19 & 67.46 & 40.40 & 46.97 & 74.17 & 79.66 \\
GLM-4-9B-Instruct & 90.45 & 89.23 & 85.07 & 84.22 & 83.10 & 77.47 & 53.80 & 41.80 & 30.87 & 35.89 & 28.28 & 22.73 & 78.55 & 77.77 \\
Olmo-3-7B-Instruct & 87.21 & 90.15 & 75.77 & 85.67 & 80.07 & 81.09 & 51.30 & 72.00 & 51.26 & 52.18 & 30.30 & 30.81 & 70.54 & 78.67 \\
GPT-OSS-20B & 63.72 & 83.25 & 70.99 & 87.71 & 67.29 & 82.69 & \textbf{96.30} & 95.30 & 61.37 & 60.24 & 46.97 & 50.51 & 18.63 & 13.92 \\
Qwen3-30B-A3B & \textbf{95.03} & \textbf{94.74} & \textbf{94.88} & \textbf{94.88} & \textbf{91.38} & \textbf{91.41} & 72.30 & 71.40 & \textbf{77.78} & \textbf{77.64} & \textbf{59.09} & \textbf{61.11} & 80.02 & \textbf{80.02} \\
\bottomrule
\end{tabular}
}
\caption{
    Performance on Reading Comprehension and Multi-Discipline Reasoning benchmarks (Part 3). Scores are reported as accuracy. The number of few-shot examples is in parentheses. The best model is in \textbf{bold}.
}
\label{tab:reasoning_part3_reading}
\end{sidewaystable*}

\iftaclpubformat

\onecolumn

\fi

\end{document}